\documentclass[journal,article,submit,moreauthors,pdftex]{Definitions/template}
\usepackage[normalem]{ulem} % For using the command \uuline{} in the \abstract 
\usepackage{amsfonts} % Use for some special math mark
\usepackage{mathcomp} % For permille mark
\usepackage{CJKutf8} % For Chinese font
\usepackage{pifont} % For some special mark 
\usepackage{bm} % Forr math enviroment bold format
\usepackage{bbm} % For some math fancy characters
\graphicspath{{./Definitions/}} % Use for import path the figures paper used

\usepackage{float} % 固定位置
\usepackage[utf8]{inputenc}
\usepackage[T1]{fontenc}
\usepackage{indentfirst} % 确保每段首行缩进
\usepackage{makecell} % 
\usepackage{amsmath}
\usepackage{seqsplit}
\usepackage[edges]{forest}
\usepackage{adjustbox}
\usepackage{tikz}
\usetikzlibrary{shapes.callouts, shadows}
\usepackage{amssymb}
\usepackage{booktabs}
\usepackage{enumitem}
\date{}

\makeatletter
\def\T@n@@nc@d@ngM@cr@M@d{}
\def\LY@n@@nc@d@ngM@cr@M@d{}
\makeatother

\let\orignewcommand\newcommand  % Store the original \newcommand
\let\newcommand\providecommand  % Make \newcommand behave like \providecommand
\usepackage{verse}
\let\newcommand\orignewcommand  % Use the original `\newcommand` in future

\newsavebox\foobox

\renewcommand{\dagger}{\mathchar"2279}
\renewcommand{\ddagger}{\mathchar"227A}

\newcommand{\mmathit}[1]{
  \ifthenelse{\equal{#1}{\ln}}{\mathit{ln}}{
    \ifthenelse{\equal{#1}{\max}}{\mathit{max}}{\mathit{#1}}
  }
}
\makeatother
\robustify{\footnote}

\DeclareUnicodeCharacter{1E45}{\.{n}}
\DeclareUnicodeCharacter{1E41}{\.{m}}
\DeclareUnicodeCharacter{2003}{\quad}
\DeclareUnicodeCharacter{2009}{\thinspace}
\DeclareUnicodeCharacter{2002}{\enspace{}}
\DeclareUnicodeCharacter{2005}{\thinspace}
\DeclareUnicodeCharacter{0263}{\textipa{G}}
\DeclareUnicodeCharacter{A0}{~}
\DeclareUnicodeCharacter{2460}{\textcircled{\scriptsize{1}}}
\DeclareUnicodeCharacter{2461}{\textcircled{\scriptsize{2}}}
\DeclareUnicodeCharacter{2462}{\textcircled{\scriptsize{3}}}
\DeclareUnicodeCharacter{2463}{\textcircled{\scriptsize{4}}}
\DeclareUnicodeCharacter{2464}{\textcircled{\scriptsize{5}}}
\DeclareUnicodeCharacter{2465}{\textcircled{\scriptsize{6}}}
\DeclareUnicodeCharacter{2466}{\textcircled{\scriptsize{7}}}
\DeclareUnicodeCharacter{2467}{\textcircled{\scriptsize{8}}}
\DeclareUnicodeCharacter{2468}{\textcircled{\scriptsize{9}}}
\DeclareUnicodeCharacter{2070}{\textsuperscript{0}}
\DeclareUnicodeCharacter{2074}{\textsuperscript{4}}
\DeclareUnicodeCharacter{2075}{\textsuperscript{5}}
\DeclareUnicodeCharacter{2076}{\textsuperscript{6}}
\DeclareUnicodeCharacter{2077}{\textsuperscript{7}}
\DeclareUnicodeCharacter{2078}{\textsuperscript{8}}
\DeclareUnicodeCharacter{2079}{\textsuperscript{9}}
\DeclareUnicodeCharacter{02C2}{<}
\DeclareUnicodeCharacter{2033}{\relax\ifmmode '' \else $''$\fi}
\DeclareUnicodeCharacter{2034}{\relax\ifmmode ''' \else $'''$\fi}
\DeclareUnicodeCharacter{2026}{\relax\ifmmode … \else $\ldots$\fi}
\DeclareUnicodeCharacter{0229}{\c{e}}
\DeclareUnicodeCharacter{016F}{\r{u}}
\DeclareUnicodeCharacter{127}{\relax\ifmmode\rm\hbar\else $\rm\hbar$\fi}
\DeclareUnicodeCharacter{3AC}{\relax\ifmmode\acute{\alpha}\else $\acute{\alpha}$\fi}
\DeclareUnicodeCharacter{3AD}{\relax\ifmmode\acute{\varepsilon}\else $\acute{\varepsilon}$\fi}
\DeclareUnicodeCharacter{3AE}{\relax\ifmmode\acute{\eta}\else $\acute{\eta}$\fi}
\DeclareUnicodeCharacter{3AF}{\relax\ifmmode\acute{\iota}\else $\acute{\iota}$\fi}
\DeclareUnicodeCharacter{3CC}{\relax\ifmmode\acute{o}\else $\acute{o}$\fi}
\DeclareUnicodeCharacter{3CD}{\relax\ifmmode\acute{\upsilon}\else $\acute{\upsilon}$\fi}
\DeclareUnicodeCharacter{3CE}{\relax\ifmmode\acute{\omega}\else $\acute{\omega}$\fi}
\DeclareUnicodeCharacter{391}{A}
\DeclareUnicodeCharacter{392}{B}
\DeclareUnicodeCharacter{395}{E}
\DeclareUnicodeCharacter{396}{Z}
\DeclareUnicodeCharacter{397}{H}
\DeclareUnicodeCharacter{399}{I}
\DeclareUnicodeCharacter{39A}{K}
\DeclareUnicodeCharacter{39C}{M}
\DeclareUnicodeCharacter{39D}{N}
\DeclareUnicodeCharacter{39F}{O}
\DeclareUnicodeCharacter{3A1}{P}
\DeclareUnicodeCharacter{3A4}{T}
\DeclareUnicodeCharacter{3A7}{X}

\DeclareUnicodeCharacter{27E6}{\relax\ifmmode \llbracket \else $\llbracket$\fi}
\DeclareUnicodeCharacter{27E7}{\relax\ifmmode \rrbracket \else $\rrbracket$\fi}

\DeclareUnicodeCharacter{1D434}{\relax\ifmmode A \else $A$\fi}
\DeclareUnicodeCharacter{1D435}{\relax\ifmmode B \else $B$\fi}
\DeclareUnicodeCharacter{1D436}{\relax\ifmmode C \else $C$\fi}
\DeclareUnicodeCharacter{1D437}{\relax\ifmmode D \else $D$\fi}
\DeclareUnicodeCharacter{1D438}{\relax\ifmmode E \else $E$\fi}
\DeclareUnicodeCharacter{1D439}{\relax\ifmmode F \else $F$\fi}
\DeclareUnicodeCharacter{1D43A}{\relax\ifmmode G \else $G$\fi}
\DeclareUnicodeCharacter{1D43B}{\relax\ifmmode H \else $H$\fi}
\DeclareUnicodeCharacter{1D43C}{\relax\ifmmode I \else $I$\fi}
\DeclareUnicodeCharacter{1D43D}{\relax\ifmmode J \else $J$\fi}
\DeclareUnicodeCharacter{1D43E}{\relax\ifmmode K \else $K$\fi}
\DeclareUnicodeCharacter{1D43F}{\relax\ifmmode L \else $L$\fi}
\DeclareUnicodeCharacter{1D440}{\relax\ifmmode M \else $M$\fi}
\DeclareUnicodeCharacter{1D441}{\relax\ifmmode N \else $N$\fi}
\DeclareUnicodeCharacter{1D442}{\relax\ifmmode O \else $O$\fi}
\DeclareUnicodeCharacter{1D443}{\relax\ifmmode P \else $P$\fi}
\DeclareUnicodeCharacter{1D444}{\relax\ifmmode Q \else $Q$\fi}
\DeclareUnicodeCharacter{1D445}{\relax\ifmmode R \else $R$\fi}
\DeclareUnicodeCharacter{1D446}{\relax\ifmmode S \else $S$\fi}
\DeclareUnicodeCharacter{1D447}{\relax\ifmmode T \else $T$\fi}
\DeclareUnicodeCharacter{1D448}{\relax\ifmmode U \else $U$\fi}
\DeclareUnicodeCharacter{1D449}{\relax\ifmmode V \else $V$\fi}
\DeclareUnicodeCharacter{1D44A}{\relax\ifmmode W \else $W$\fi}
\DeclareUnicodeCharacter{1D44B}{\relax\ifmmode X \else $X$\fi}
\DeclareUnicodeCharacter{1D44C}{\relax\ifmmode Y \else $Y$\fi}
\DeclareUnicodeCharacter{1D44D}{\relax\ifmmode Z \else $Z$\fi}
\DeclareUnicodeCharacter{1D44E}{\relax\ifmmode a \else $a$\fi}
\DeclareUnicodeCharacter{1D44F}{\relax\ifmmode b \else $b$\fi}
\DeclareUnicodeCharacter{1D450}{\relax\ifmmode c \else $c$\fi}
\DeclareUnicodeCharacter{1D451}{\relax\ifmmode d \else $d$\fi}
\DeclareUnicodeCharacter{1D452}{\relax\ifmmode e \else $e$\fi}
\DeclareUnicodeCharacter{1D453}{\relax\ifmmode f \else $f$\fi}
\DeclareUnicodeCharacter{1D454}{\relax\ifmmode g \else $g$\fi}
\DeclareUnicodeCharacter{1D456}{\relax\ifmmode i \else $i$\fi}
\DeclareUnicodeCharacter{1D457}{\relax\ifmmode j \else $j$\fi}
\DeclareUnicodeCharacter{1D458}{\relax\ifmmode k \else $k$\fi}
\DeclareUnicodeCharacter{1D459}{\relax\ifmmode l \else $l$\fi}
\DeclareUnicodeCharacter{1D45A}{\relax\ifmmode m \else $m$\fi}
\DeclareUnicodeCharacter{1D45B}{\relax\ifmmode n \else $n$\fi}
\DeclareUnicodeCharacter{1D45C}{\relax\ifmmode o \else $o$\fi}
\DeclareUnicodeCharacter{1D45D}{\relax\ifmmode p \else $p$\fi}
\DeclareUnicodeCharacter{1D45E}{\relax\ifmmode q \else $q$\fi}
\DeclareUnicodeCharacter{1D45F}{\relax\ifmmode r \else $r$\fi}
\DeclareUnicodeCharacter{1D460}{\relax\ifmmode s \else $s$\fi}
\DeclareUnicodeCharacter{1D461}{\relax\ifmmode t \else $t$\fi}
\DeclareUnicodeCharacter{1D462}{\relax\ifmmode u \else $u$\fi}
\DeclareUnicodeCharacter{1D463}{\relax\ifmmode v \else $v$\fi}
\DeclareUnicodeCharacter{1D464}{\relax\ifmmode w \else $w$\fi}
\DeclareUnicodeCharacter{1D465}{\relax\ifmmode x \else $x$\fi}
\DeclareUnicodeCharacter{1D466}{\relax\ifmmode y \else $y$\fi}
\DeclareUnicodeCharacter{1D467}{\relax\ifmmode z \else $z$\fi}

\DeclareUnicodeCharacter{1E67}{\.{\v s}}
\DeclareUnicodeCharacter{1E11}{\relax\ifmmode \c{d} \else $\c{d}$\fi}
\DeclareUnicodeCharacter{1ECB}{\relax\ifmmode \d{i} \else $\d{i}$\fi}
\DeclareUnicodeCharacter{1D8D}{\relax\ifmmode \textlhookx \else $\textlhookx$\fi}
\DeclareUnicodeCharacter{104}{\relax\ifmmode \k{A} \else $\k{A}$\fi}
\DeclareUnicodeCharacter{211E}{\relax\ifmmode \textrecipe \else $\textrecipe$\fi}
\DeclareUnicodeCharacter{29D}{\relax\ifmmode \textctj \else $\textctj$\fi}

\DeclareUnicodeCharacter{1E2E}{\'{\"I}}
\DeclareUnicodeCharacter{23F}{\textrts}

\DeclareUnicodeCharacter{2C73}{\varw}

\DeclareUnicodeCharacter{2127}{\mho}

\DeclareUnicodeCharacter{28C}{\textturnv}
\DeclareUnicodeCharacter{252}{\textturnscripta}
\DeclareUnicodeCharacter{259}{\schwa}
\DeclareUnicodeCharacter{25B}{\m{e}}
\DeclareUnicodeCharacter{266}{\m{h}}
\DeclareUnicodeCharacter{127}{\B{h}}
\DeclareUnicodeCharacter{27E}{\textfishhookr}
\DeclareUnicodeCharacter{281}{\textinvscr}

\continuouspages{yes}
\firstpage{1} 
\pubvolume{1}
\issuenum{1}
\articlenumber{12345}
\pubyear{2025}
\copyrightyear{2025}
\datereceived{Day Month Year}
\dateaccepted{Day Month Year}
\dateonlinefirst{}
\datepublished{}

\Title{The Rise of Sparse Mixture-of-Experts:A Survey from Algorithmic Foundations to Decentralized Architectures and Vertical Domain Applications}
\Author{Dong Pan\textsuperscript{1}, Bingtao Li\textsuperscript{1}, Yongsheng Zheng\textsuperscript{1}, Jiren Ma\textsuperscript{1}, Victor Fei\textsuperscript{2}}

\AuthorNames{Dong Pan, Bingtao Li, Yongsheng Zheng, Jiren Ma, Victor Fei}

\address{%
\textsuperscript{1}FEDIMOSS TECH HK LIMITED, Hong Kong\\
\textsuperscript{2}Ormi Labs, Inc., California, USA
}%`

\corres{Corresponding Author: Bingtao Li. Email: bingtaoli@fedimoss.com}

\abstract{The sparse Mixture of Experts (MoE) architecture has evolved as a powerful approach for scaling deep learning models to more parameters with comparable computation cost. As an important branch of large language model (LLM), MoE model only activate a subset of experts based on a routing network. This sparse conditional computation mechanism significantly improves computational efficiency, paving a promising path for greater scalability and cost-efficiency. It not only enhance downstream applications such as natural language processing, computer vision, and multimodal in various horizontal domains, but also exhibit broad applicability across vertical domains including medical diagnosis, autonomous driving, financial analysis, and business intelligence. Despite the growing popularity and application of MoE models across various domains, there lacks a systematic exploration of recent advancements of MoE in many important fields. Existing surveys on MoE suffer from limitations such as lack coverage or none extensively exploration of key areas. This survey seeks to fill these gaps. In this paper, Firstly, we examine the foundational principles of MoE, with an in-depth exploration of its core components—the routing network and expert network. Subsequently, we extend beyond the centralized paradigm to the decentralized paradigm, which unlocks the immense untapped potential of decentralized infrastructure, enables democratization of MoE development for broader communities, and delivers greater scalability and cost-efficiency. Furthermore we focus on exploring its vertical domain applications. Finally, we also identify key challenges and promising future research directions. To the best of our knowledge, this survey is currently the most comprehensive review in the field of MoE. We aim for this article to serve as a valuable resource for both researchers and practitioners, enabling them to navigate and stay up-to-date with the latest advancements.}


\definecolor{line-color}{RGB}{0, 120, 180}
\definecolor{fill-color}{RGB}{0, 0, 255}

\tikzstyle{category}=[
    rectangle,
    draw=line-color,
    rounded corners,
    text opacity=1,
    minimum height=1.5em,
    minimum width=5em,
    inner sep=2pt,
    align=center,
    fill opacity=.5,
]
\tikzstyle{leaf}=[category,minimum height=1.5em,
fill=fill-color!60, text width=20em,  text=black,align=left,font=\small,
inner xsep=2pt,
inner ysep=1pt,
]

\begin{document}
\section{Introduction} \label{sect:s1}
Recent advances in artificial intelligence (AI), especially regarding large language model (LLM), predominantly stem from scaling principles\cite{kaplan2020scaling,brown2020language,achiam2023gpt} that larger model sizes bring better model quality—a phenomenon formally described as the Scaling Law. Despite its oversimplified nature, this fundamental principle continues to steer AI research evolution. However, extreme-scale model expansion also incurs extremely high computational costs.\\
\indent To this end, architectures based on sparse Mixture of Experts (MoE) \cite{shazeer2017outrageously,lepikhin2020gshard,fedus2022switch,zoph2022st,du2022glam,dai2024deepseekmoe} have paved a promising path, enabling the scaling of foundation models to larger sizes at comparable computational cost. A recent open-source MoE model\cite{liu2024deepseek} which integrates sparsely-gated MoE with Transformer-based foundation models\cite{vaswani2017attention,radford2018improving}, has surpassed other open-source alternatives and demonstrated performance comparable to prominent closed-source models such as GPT-4o\cite{hurst2024gpt}, thereby unlocking the broader application potential of this three-decade-old technology\cite{jacobs1991adaptive}.\\
\indent Beyond efficiency advantages, the MoE architectures also offer opportunities to enhance model interpretability \cite{mustafa2022multimodal,akrour2021continuous,pavlitska2023sparsely,mixtureofexperts,zoph2022st}. By learning its intrinsic allocation mechanism, researchers can gain insights into how different "experts" specialize in handling specific types of data or tasks. This interpretability not only deepens our understanding of model behavior but also paves new pathways for designing more robust and transparent AI systems.\\
\indent However, some recent MoE models\cite{KimiK2,yang2025qwen3} have scaled to 1T parameters with context windows exceeding 128K tokens, dramatically increasing computational resource demands. This exponential growth poses significant challenges. High-performance computing clusters for advanced MoE research and development remain unaffordable for resource-limited individual researchers and small laboratories. Only a handful of large corporations and institutions possess sufficient resources to develop such models, creating quasi-monopolies that stifle AI innovation.\\
\indent Urgent adoption of efficient training and inference paradigms is imperative for sustainable scaling. Current advanced frameworks\cite{rajbhandari2022deepspeed,nie2022hetumoe,gale2023megablocks,hwang2023tutel,xue2024openmoe,deepep2025} primarily utilize limited homogeneous resources, operating under centralized paradigm. Crucially, decentralized clusters and consumer-level devices, usually have a significant amount of computing resources than centralized clusters. Despite harboring immense untapped potential, these computing resources are typically overlooked due to lower bandwidth and compute power. Decentralized paradigm\cite{ryabinin2020towards,mcallister2025decentralized} emerges as a promising solution. This distributed paradigm integrates heterogeneous resources across individual consumer GPUs, and clusters, fully utilizes the computing resources, enables democratization of MoE development for broader communities, delivers greater scalability and cost-efficiency.\\
\subsection{\textbf{Related Work}}
\indent While several related surveys predate this work, notable gaps still remain. For instance, among the comprehensive studies,\cite{yuksel2012twenty} covers pre-deeplearning developments, \cite{fedus2022review} omits recent breakthroughs in the field, \cite{cai2025survey,mu2025comprehensive} lack coverage of decentralized MoE paradigms, and none extensively explore MoE applications in vertical industries. Furthermore, there are focus-limited studies, \cite{gan2025mixture} focuses on big data applications, \cite{liu2024survey} focuses on inference acceleration, \cite{xu2025decentralization} focuses on applications of wireless communication scenarios. Our survey bridges these gaps by conducting an in-depth exploration of MoE architectures in both centralized and decentralized infrastructure, scenario-specific applications in critical vertical domains. \tabref{tabref:comparison} highlights the differences.\\
\begin{table}[H]
    \centering
    \caption{Comparison of our survey with related surveys.}
    \resizebox{\textwidth}{!}{
    \begin{tabular}{c|c|c|c|c|c|c|c|c} 
     \toprule
     \textbf{Surveys} & \textbf{\cite{yuksel2012twenty}} & \textbf{\cite{fedus2022review}} & \textbf{\cite{cai2025survey}} & \textbf{\cite{mu2025comprehensive}} & \textbf{\cite{gan2025mixture}} & \textbf{\cite{liu2024survey}} & \textbf{\cite{xu2025decentralization}} & \textbf{Ours} \\ [1ex] 
     \midrule     
     \makecell[c]{\textbf{Comprehensive introduction of MoE} \\ core designs and recent advancements} & \makecell[c]{$\times$} & \makecell[c]{$\times$} & \makecell[c]{$\checkmark$} & \makecell[c]{$\checkmark$} & \makecell[c]{$\times$} & \makecell[c]{$\times$} & \makecell[c]{$\checkmark$} & \makecell[l]{$\checkmark$} \\
     \midrule
     \makecell[c]{\textbf{Delineation of the decentralized architecture paradigm}} & \makecell[c]{$\times$} & \makecell[c]{$\times$} & \makecell[c]{$\times$} & \makecell[c]{$\times$} & \makecell[c]{$\times$} & \makecell[c]{$\times$} & \makecell[c]{$\times$} & \makecell[l]{$\checkmark$} \\  
     \midrule
     \makecell[c]{\textbf{Extensively exploration of vertical domain applications}} & \makecell[c]{$\times$} & \makecell[c]{$\times$} & \makecell[c]{$\times$} & \makecell[c]{$\times$} & \makecell[c]{$\checkmark$} & \makecell[c]{$\times$} & \makecell[c]{$\times$} & \makecell[l]{$\checkmark$} \\ [1ex]  
     \bottomrule     
    \end{tabular}
    }  
    \label{tabref:comparison}
\end{table}

\subsection{Contributions}

\begin{itemize}
    \item \textbf{Comprehensive and Timely Survey:} We present the most comprehensive review in the field of MoE, systematically identifying studies from algorithmic foundations to decentralized architectures and vertical domain applications. Our survey meticulously analyzes relevant research, examining their motivations, technical principles, and key factors requiring consideration within them, offering a valuable reference for researchers and practitioners, tracking the latest research developments and inspiring new ideas in this explosively evolving field.
    \item \textbf{In-Depth Exploration of Decentralized Architecture Paradigm:} We delve into decentralized paradigm, which unlocks the immense untapped potential of decentralized infrastructure. Our survey delineates critical challenges faced by the decentralized paradigm and reviews existing related research efforts addressing these issues, moving beyond traditional centralized paradigm toward enabling democratization of MoE development for broader communities, paving the way for greater scalability and cost-efficiency. This evolves the development paradigm for MoE models from "competing for resources" to "on-demand scalability."
    \item \textbf{Diverse Applications of Vertical Domains:} We extensively explore applications of MoE in typical vertical domains, providing an overall understanding of how MoE can be applied to specific tasks.    
\end{itemize}

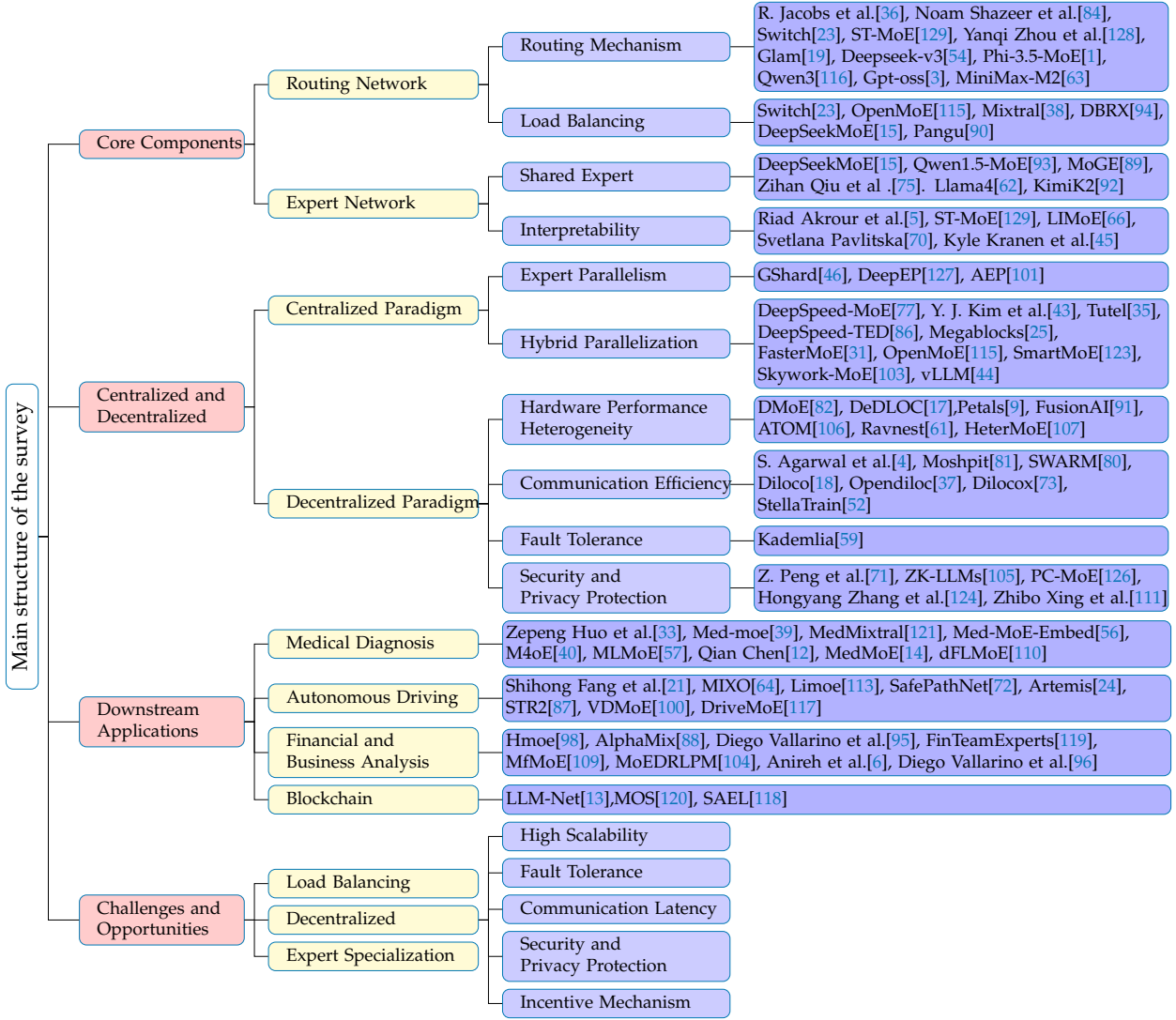
\begin{figure}[H]
  \centering    
  \begin{adjustbox}{max width=\textwidth, center}
  \begin{forest}
  forked edges,
  for tree={
  grow=east,
  reversed=true,%increase counter-clockwise
  anchor=base west,
  parent anchor=east,
  child anchor=west,
  base=left,
  font=\large,
  rectangle,
  draw=line-color,
  rounded corners,align=left,
  minimum width=2.5em,
s sep=4pt,
inner xsep=10pt,
inner ysep=2pt,
align=left,
ver/.style={rotate=90, child anchor=north, parent anchor=south, anchor=center},
  },
  where level=1{text width=6.6em,font=\small,}{},
  where level=2{text width=8.9em,font=\small}{},  
  where level=3{text width=9.8em,font=\small}{},
  [Main structure of the survey,ver,fill=white!20 
    [Core Components,fill=red!20
      [Routing Network,fill=yellow!20
      [Routing Mechanism,fill=blue!20
      [R. Jacobs et al.\cite{jacobs1991adaptive}{,} Noam Shazeer et al.\cite{shazeer2017outrageously}{,}\\Switch\cite{fedus2022switch}{,} ST-MoE\cite{zoph2022st}{,} Yanqi Zhou et al.\cite{zhou2202mixture}{,}\\Glam\cite{du2022glam}{,} Deepseek-v3\cite{liu2024deepseek}{,} Phi-3.5-MoE\cite{abdin2024phi}{,}\\Qwen3\cite{yang2025qwen3}{,} Gpt-oss\cite{agarwal2025gpt}{,} MiniMax-M2\cite{MiniMax-M2},leaf,text width=20.8em]
      ]      
      [Load Balancing,fill=blue!20
      [Switch\cite{fedus2022switch}{,} OpenMoE\cite{xue2024openmoe}{,} Mixtral\cite{jiang2024mixtral}{,} DBRX\cite{DBRX}{,} \\DeepSeekMoE\cite{dai2024deepseekmoe}{,} Pangu\cite{tang2025ultrapangu},leaf,text width=20.8em]
      ]      
      ]
      [Expert Network,fill=yellow!20
      [Shared Expert,fill=blue!20
      [DeepSeekMoE\cite{dai2024deepseekmoe}{,} Qwen1.5-MoE\cite{qwen_moe}{,} MoGE\cite{tang2025pangu}{,}\\Zihan Qiu et al .\cite{qiu2025demonsdetailimplementingload}{.} Llama4\cite{llamamoe}{,} KimiK2\cite{KimiK2}, leaf,text width=20.8em]
      ]      
      [Interpretability,fill=blue!20
      [Riad Akrour et al.\cite{akrour2021continuous}{,} ST-MoE\cite{zoph2022st}{,} LIMoE\cite{mustafa2022multimodal}{,}\\Svetlana Pavlitska\cite{pavlitska2023sparsely}{,} Kyle Kranen et al.\cite{mixtureofexperts},leaf,text width=20.8em]
      ]           
      ]      
    ]
    [Centralized and \\Decentralized,fill=red!20
      [Centralized Paradigm,fill=yellow!20
      [Expert Parallelism,fill=blue!20
      [GShard\cite{lepikhin2020gshard}{,} DeepEP\cite{deepep2025}{,} AEP\cite{wang2025toward},leaf,text width=20.8em]
      ]
      [Hybrid Parallelization,fill=blue!20
      [DeepSpeed-MoE\cite{rajbhandari2022deepspeed}{,} Y. J. Kim et al.\cite{kim2021scalable}{,} Tutel\cite{hwang2023tutel}{,}\\DeepSpeed-TED\cite{singh2023hybrid}{,} Megablocks\cite{gale2023megablocks}{,}\\FasterMoE\cite{he2022fastermoe}{,} OpenMoE\cite{xue2024openmoe}{,} SmartMoE\cite{zhai2023smartmoe}{,}\\Skywork-MoE\cite{wei2024skyworkmoedeepdivetraining}{,} vLLM\cite{10.1145/3600006.3613165},leaf,text width=20.8em]
      ]
      ]
      [Decentralized Paradigm,fill=yellow!20
      [Hardware Performance \\Heterogeneity,fill=blue!20
      [DMoE\cite{ryabinin2020towards}{,} DeDLOC\cite{diskin2021distributed}{,}Petals\cite{borzunov2022petals}{,} FusionAI\cite{tang2023fusionai}{,}\\
     ATOM\cite{wu2024atom}{,} Ravnest\cite{menon2024ravnest}{,} HeterMoE\cite{wu2025hetermoe},leaf,text width=20.8em]
      ]
      [Communication Efficiency,fill=blue!20
      [S. Agarwal et al.\cite{agarwal2022utility}{,} Moshpit\cite{ryabinin2021moshpit}{,} SWARM\cite{ryabinin2023swarm}{,}\\
     Diloco\cite{douillard2023diloco}{,} Opendiloc\cite{jaghouar2024opendiloco}{,} Dilocox\cite{qi2025dilocox}{,}\\StellaTrain\cite{lim2024accelerating},leaf,text width=20.8em]
      ]
      [Fault Tolerance,fill=blue!20
      [Kademlia\cite{maymounkov2002kademlia},leaf,text width=20.8em]
      ]
      [Security and \\Privacy Protection,fill=blue!20
      [Z. Peng et al.\cite{peng2025survey}{,} ZK-LLMs\cite{wellington2024basedai}{,} PC-MoE\cite{zhang2025memory}{,}\\
     Hongyang Zhang et al.\cite{zhang2024towards}{,} Zhibo Xing et al.\cite{xing2025zero},leaf,text width=20.8em]
      ]
      ]    
    ]
    [Downstream \\Applications,fill=red!20
      [Medical Diagnosis,fill=yellow!20
      [Zepeng Huo et al.\cite{huo2021sparse}{,} Med-moe\cite{jiang2024med}{,} MedMixtral\cite{yuan2024efficient}{,} Med-MoE-Embed\cite{liu2024medical-e}{,}\\M4oE\cite{jiang2024m4oe}{,} MLMoE\cite{liu2024medical}{,} Qian Chen\cite{chen2024low}{,} MedMoE\cite{chopra2025medmoe}{,} dFLMoE\cite{xie2025dflmoe},leaf,text width=33.8em]
      ]
      [Autonomous Driving,fill=yellow!20
      [Shihong Fang et al.\cite{fang2020multi}{,} MIXO\cite{morra2023mixo}{,} Limoe\cite{xu2025limoe}{,} SafePathNet\cite{pini2023safe}{,} Artemis\cite{feng2025artemis}{,}\\STR2\cite{sun2024generalizing}{,} VDMoE\cite{wang2024efficient}{,} DriveMoE\cite{yang2025drivemoe},leaf,text width=33.8em]
      ]
      [Financial and \\Business Analysis,fill=yellow!20
      [Hmoe\cite{wang2024hmoe}{,} AlphaMix\cite{sun2022quantitative}{,} Diego Vallarino et al.\cite{vallarino2024dynamic}{,} FinTeamExperts\cite{yu2024finteamexperts}{,}\\ 
MfMoE\cite{xiao2024autoeis}{,} MoEDRLPM\cite{wei2025deep}{,} Anireh et al.\cite{anireh2024model}{,} Diego Vallarino et al.\cite{vallarino2025consumers},leaf,text width=33.8em]
      ]
      [Blockchain,fill=yellow!20
      [LLM-Net\cite{chong2025llm}{,}MOS\cite{yuan2025mos}{,} SAEL\cite{yu2025saelleveraginglargelanguage},leaf,text width=33.8em]
      ]      
    ]
    [Challenges and \\Opportunities,fill=red!20
    [Load Balancing,fill=yellow!20]
    [Decentralized,fill=yellow!20
    [High Scalability,fill=blue!20]
    [Fault Tolerance,fill=blue!20]
    [Communication Latency,fill=blue!20]
    [Security and \\Privacy Protection,fill=blue!20]
    [Incentive Mechanism,fill=blue!20]
    ]
    [Expert Specialization,fill=yellow!20]
    ]  
  ]
  \end{forest}
  \end{adjustbox}  
  \caption{Main structure of the survey.}
  \label{fig:survey}
\end{figure}

\indent The remainder of this survey is organized as follows:
Section 2 comprehensively introduces the core component design of sparse MoE models and examines various factors requiring consideration in these components.
Section 3 extends beyond the centralized paradigm to delve into the decentralized paradigm, unlocking their potential for collaborative training and inference of MoE models in heterogeneous environments.
Section 4 shifts focus from widely studied horizontal applications (e.g., NLP, CV, Multimodal) to vertical industry applications, such as medical diagnosis, autonomous driving, finance analysis, business intelligence and blockchain. This aims to provide readers with a comprehensive reference for implementing MoE in specific tasks.
Section 5 analyzes critical challenges and emerging opportunities.
Section 6 concludes this paper.
The overall structure of this survey is illustrated in \fig{fig:survey}:\\
\section{Core Components of MoE} \label{sect:s2}
The concept of MoE originated from the 1991 Adaptive Mixture of Local Experts\cite{jacobs1991adaptive}. MoE model introduces the idea of experts, dividing the network into multiple independent subnetworks (called experts). MoE model dynamically determines which experts should process which tokens, typically activating only the most relevant few experts for each token. The remaining experts do not participate in the computation, significantly improving computational efficiency. This allows for a significant increase in model parameter size under the same computational budget\cite{deanintroducing,yuksel2012twenty,fedus2022review}. \tabref{tabref:moemodels} summarizes the key characteristics of state-of-the-art open-source MoE models.\\
\indent The MoE model primarily consists of two key components: 
\begin{itemize}
    \item \textbf{Routing Network:} This component determines which tokens are sent to which experts. The routing mechanism is a critical aspect of MoE systems. Routing network is composed of learnable parameters and is trained alongside the rest of the network during pre-training.
    \item \textbf{Expert Network:} By splitting the dense feed-forward neural network (FFN) layers into multiple independent parts, each of which is an independent neural network—expert network. In practice, these experts are typically FFN units, but they can also be CNN or more complex network structures.     
\end{itemize}

\begin{table}[H] 
 \tablesize{\footnotesize}
\caption{A Summary of State-of-the-Art Open-Source MoE Models.}
\label{tabref:moemodels}
\newcolumntype{C}{>{\centering\arraybackslash}X}
\begin{tabularx}{\textwidth}{CCCCCC}
\toprule
\textbf{Model}	&\textbf{Affiliation} &\textbf{\makecell{Expert Count \\ (Routed + Shared)}} &\textbf{\makecell{Activation Count \\ (Routed + Shared)}} &\textbf{\makecell{Model Params \\ (Activation/Total)}} &\textbf{Time}\\
\midrule
DeepSeekMoE\cite{dai2024deepseekmoe}	&DeepSeek	&64+2	&6+2	&2.8B/16.4B	&2024.01\\
DBRX\cite{DBRX}	&Databricks	&16+0	&4+0	&36B/132B	&2024.03\\
Mixtral\cite{jiang2024mixtral}	&Mistral AI	&8+0	&2+0	&39B/141B	&2024.04\\
Phi-3.5-MoE\cite{abdin2024phi}	&Microsoft	&16+0	&2+0	&6.6B/42B	&2024.08\\
Deepseek-v3\cite{liu2024deepseek}	&DeepSeek	&256+1	&8+1	&37B/671B	&2024.12\\
Llama4\cite{llamamoe}	&Meta	&\makecell{16+1 \\ 128+1}	&\makecell{1+1 \\ 1+1}	&\makecell{17B/109B \\ 17B/400B}	&2025.04\\
Qwen3\cite{yang2025qwen3}	&Alibaba	&\makecell{128+0 \\ 128+0}	&\makecell{8+0 \\ 8+0}	&\makecell{3B/30B \\ 22B/235B}	&2025.05\\
Pangu Ultra MoE\cite{tang2025ultrapangu}	&Huawei	&256+1	&8+1	&39B/718B	&2025.05\\
KimiK2\cite{KimiK2}	&Moonshot	&384+1 &8+1	&32B/1T	&2025.07\\
Gpt-oss\cite{agarwal2025gpt}	&OpenAI	&\makecell{32+0 \\ 128+0}	&\makecell{4+0 \\ 4+0}	&\makecell{3.6B/21B \\ 5.1B/117B}	&2025.08\\
Qwen3-Next\cite{yang2025qwen3}	&Alibaba	&512+1	&10+1	&3B/80B	&2025.09\\
MiniMax-M2\cite{MiniMax-M2}	&MiniMax	&256+0	&8+0	&10B/229B	&2025.10\\
\bottomrule
\end{tabularx}
\end{table}

\indent In the following section, we will delve into the core components of the MoE model, introduce the basic principles of key algorithms, and discuss various factors that need to be considered within them.

\subsection{Routing Network} \label{sect:s2dot1}
The routing network, also called the router, which is used to determine which tokens are route to which experts. The sparse gating functions activate a selected subset of experts or tokens, which can be considered as a form of conditional computation. 

\subsubsection{Routing Mechanism} \label{sect:s2dot1dot1}
\begin{figure}[H]
    \centering
    \includegraphics[width=1.0\linewidth]{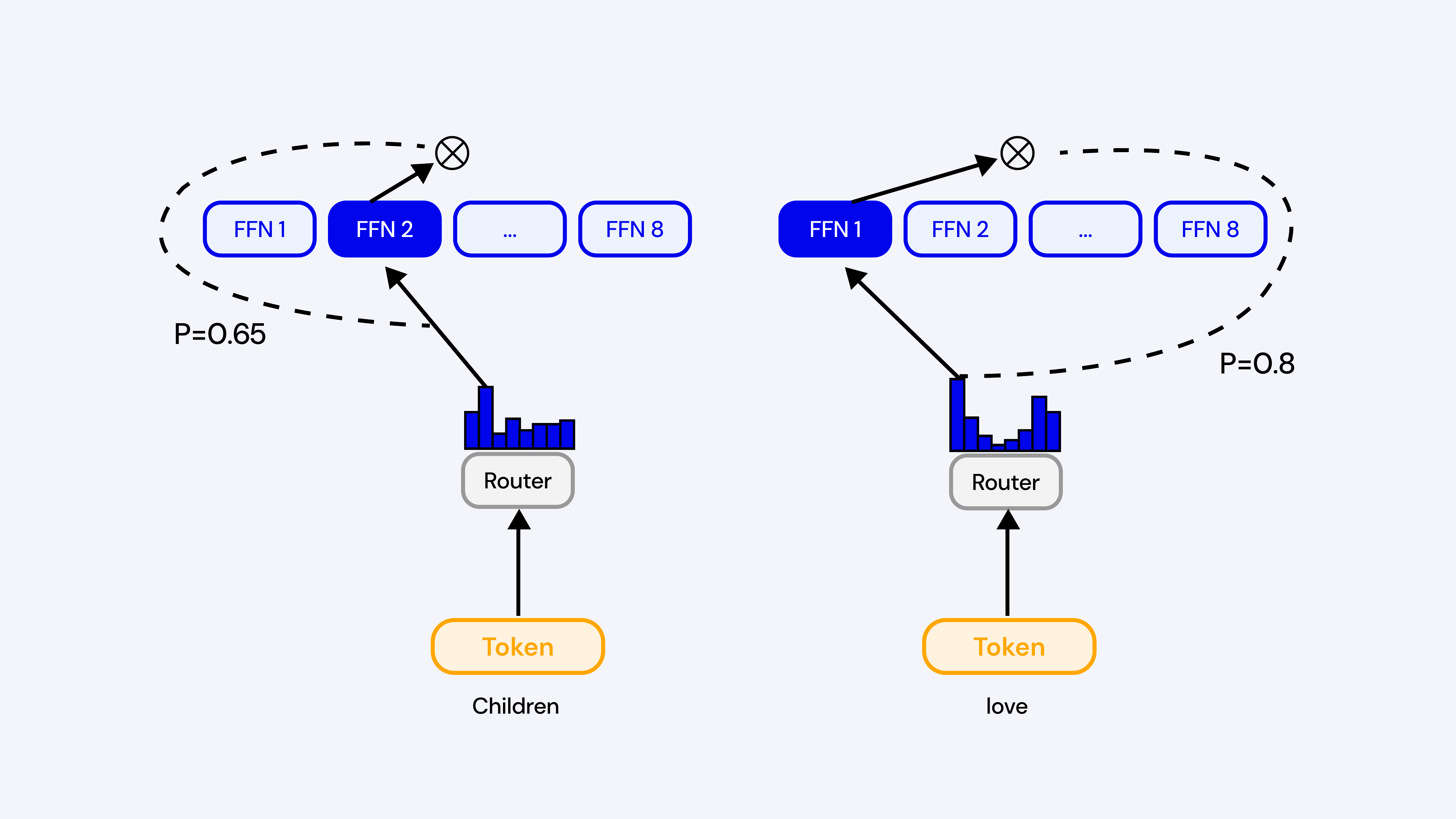}
    \caption{Illustration of token choice routing.}
    \label{fig:tokenchoice}
\end{figure}

The routing mechanism have been studied extensively\cite{shazeer2017outrageously,lepikhin2020gshard,fedus2022switch,du2022glam,liu2024deepseek,yang2025qwen3}. In this section, we provide an in-depth review of two common categories:(1) token choice routing, where each token selects the best top-k experts, and (2) expert choice routing, where each expert picks the top-k tokens.\\
\indent \cite{shazeer2017outrageously} introduces a token choice routing mechanism. In this routing mechanism, only the top-k experts determined by the router are forward-passed, while the computations of the other experts are skipped, achieving sparse activation of experts, as shown in \fig{fig:tokenchoice}. The computation process is as follows in the formula below:
\begin{equation}
\mathrm{MoE(x)} = {\sum_{i = 1}^{ \mathrm{n}}}{{ \mathrm{(G}}_i} \mathrm{(x)}{{ \mathrm{E}}_i} \mathrm{(x))}
\end{equation}

\begin{equation}
G\left({\mathbf{x}}\right)\mathrm{ = Softmax}\left({\mathrm{TopK}\left ({H\left({x}\right)\mathrm{, }k}\right)}\right)  
\end{equation}

\begin{equation}
H\left({\mathbf{x}}\right)\mathrm{ = }\left(\mathbf{x}.W_g\right)_i{\mathrm{+}}{\mathrm{StandardNormal}}\left(\right){\mathrm{.}}{\mathrm{Softplus}\left ({\left(\mathbf{x}{.W_{\mathrm{noise}}}\right)_i}\right)}  
\end{equation}

\begin{equation}
\mathrm{TopK}\left( v,k \right) = 
\begin{cases} 
v_i, & \text{if } v_i \text{ belongs to the top } k \text{ of } v, \\
\\
-\infty, & \text{otherwise}.
\end{cases}
\end{equation}

\indent where the router \( G({\mathbf{x}}) \) is obtained by multiplying the input x with a learnable weight matrix and Gaussian noise, followed by applying the Softmax function, k represents the number of experts activated per token. \( TopK({v,k}) \) refers to selecting the top k largest elements from the vector v, while assign the rest to \( {-\infty} \). The Softmax output will result in probabilities of 0 for these \( {-\infty} \) elements. Finally, only the top-k experts are selected for each token.

\begin{figure}[H]
    \centering
    \includegraphics[width=1.0\linewidth]{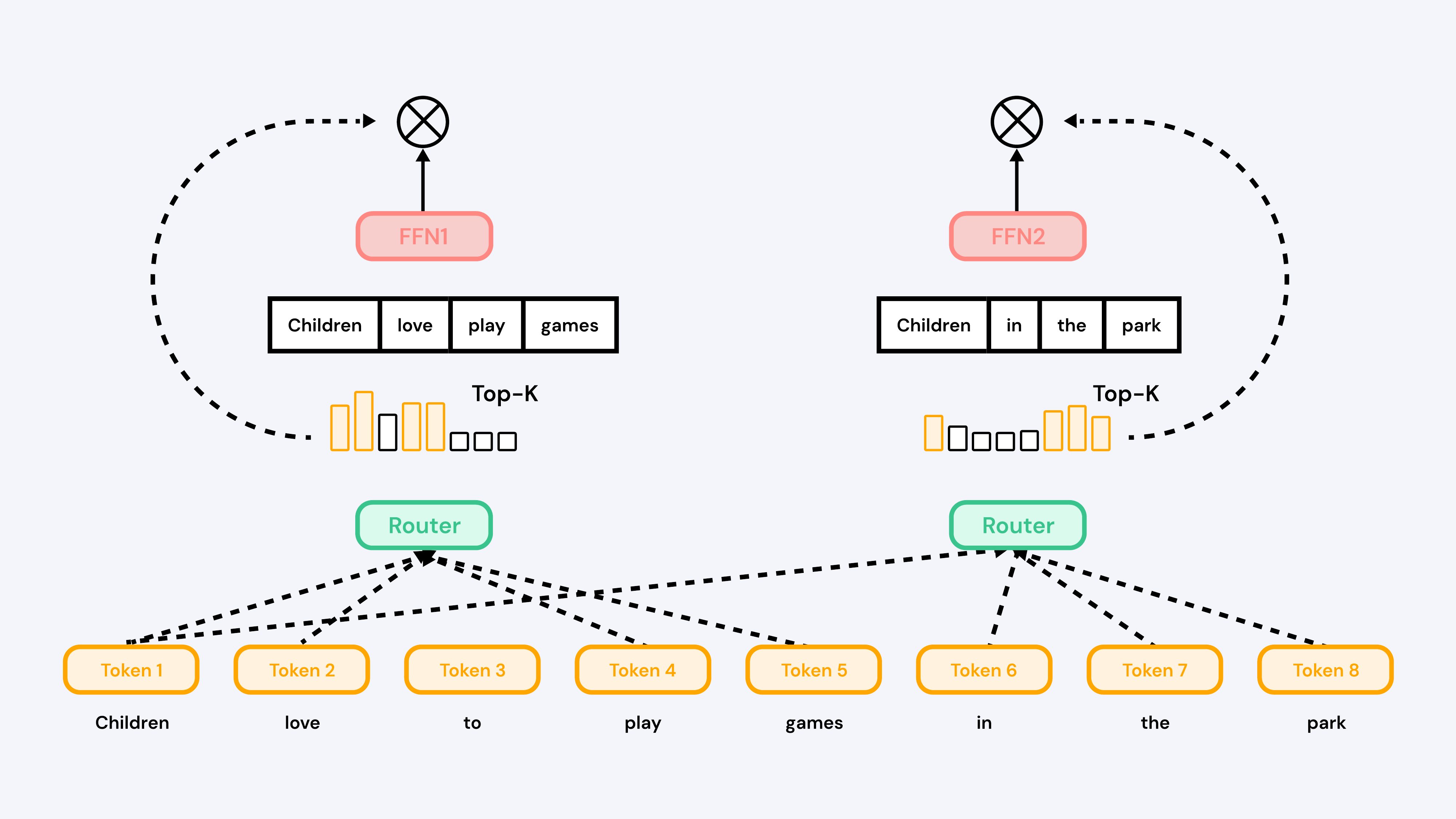}
    \caption{Illustration of expert choice routing.}
    \label{fig:expertchoice}
\end{figure}

\indent The token choice routing mechanism allows each token to select the top-k experts. However, these independent token-level decisions often lead to expert load imbalance, resulting in reduced training efficiency and suboptimal model convergence. Despite previous works add an auxiliary loss on load balancing to mitigate these issues, this auxiliary loss does not guarantee a balanced load and may potentially degrade model performance.\\
\indent \cite{zhou2202mixture} proposes a new routing method for sparsely activated MoE model, termed expert choice routing. The key idea of the routing method is to have experts pick the top-k tokens rather than tokens selecting top-k experts, as shown in \fig{fig:expertchoice}. Crucially, it allows tokens to be processed by varying numbers of experts while achieving perfect load balancing. This design simultaneously resolves two key limitations: (1) the expert-token assignment imbalance that leaves some experts under-optimized, and (2) the uniform expert count per token that ignores varying task relevance across tokens. Consequently, it enhances both training efficiency and downstream task performance.

\subsubsection{Load Balancing} \label{sect:s2dot1dot2}
In sparse MoE models, load balancing is a critical issue. Since only a small number of experts are activated, some experts may become overloaded with too many tokens, while others remain underutilized. This imbalance can lead to wasted computational resources and even cause routing collapse, ie.,all tokens are routed to a few experts, leaving others undertrained and ineffective. If experts are distributed across multiple devices, load imbalance can exacerbate computational bottlenecks.
\begin{itemize}
    \item \textbf{Auxiliary Loss:} To mitigate this issue, Expert-Level Balance Loss and Device-Level Balance Loss as auxiliary losses \cite{lepikhin2020gshard,fedus2022switch,xue2024openmoe,jiang2024mixtral,tang2025ultrapangu} are added to the total model loss. These losses encourage tokens to be evenly distributed among experts, addressing expert utilization imbalance and cross-device computational load imbalance, respectively. However, this strategy may negatively impact model performance, requiring tuning the hyperparameter \( {\alpha} \) to control the proportion of auxiliary loss in the total model loss, trading off load balancing and performance. The model's total training loss function is defined as:\\
\begin{equation}
{L_{\mathrm{total}}} = {L_{moe}}{{ + \alpha L}_{aux}}
\end{equation}
Where \( {L_{moe}} \) represents the loss function of the standard MoE, such as cross-entropy loss. In the calculation formula of \( {L_{total}} \), the hyperparameter \( {\alpha} \) is used to control the weight of the auxiliary loss in the overall model loss. Notably, a large value degrades the model's performance, while a small value leads to training instability and resource underutilization.\\
As previously mentioned, if all tokens are sent to only a few popular experts, the training efficiency will decrease. In standard MoE training, the routing network tends to primarily activate the same few experts, a tendency that can become self-reinforcing, as popular experts train faster and thus are more likely to be selected. To alleviate this issue, an expert-level auxiliary loss\cite{lepikhin2020gshard,fedus2022switch} is introduced to encourage assigning equal importance to all experts. This loss ensures that all experts receive approximately the same number of training samples, thereby balancing the selection among experts, as detailed below:\\
\begin{equation}
    P_i =\frac{1}{T}{\sum_{t = 1}^T}{s_{i,t}}
\end{equation}
\begin{equation}
    f_i =\frac{1}{K_rT}{\sum_{t = 1}^T}1 \left ({{s_{i,t}} \in Topk \left ({{s_{j,t}}|1 \leq j \leq N_r, K_r} \right )} \right )
\end{equation}
\begin{equation}
    {L_{aux}} = \alpha N_r{\sum_{i = 1}^{N_r}}f_iP_i
\end{equation}
Where \( {s_{i,t}} \) denotes the routing probability of expert i for token t forward the gating network. \( {P_{i}} \) represents the average routing probability of expert i on a set of T input tokens. If the \( {N_{r}} \) experts are load-balanced, then the \( {P_{i}} \) for each expert should be \( \frac{1}{N_{r}} \). \( {f_{i}} \) represents the probability that an arbitrary token is routed to expert i. If the \( {N_{r}} \) experts are load-balanced, then the \( {f_{i}} \) for each expert should be \( \frac{1}{N_{r}} \). \\ 
For the auxiliary loss \( {L_{aux}} \), the larger the differences in probabilities across experts, the larger the value of \( {L_{aux}} \). Conversely, the smaller the differences, the smaller \( {L_{aux}} \) becomes. The minimum value is achieved only when \( f_i\mathrm{=}\frac{1}{N_r} \) and \( P_i\mathrm{=}\frac{1}{N_r} \), indicating perfect load balancing. \\
In a distributed environment, MoE models distribute experts across multiple devices. If certain devices frequently activate experts while others rarely use their experts, it can lead to imbalanced computation load, affecting training efficiency. \cite{dai2024deepseekmoe} further introduces a device-level auxiliary loss to encourage assigning equal computation load to all devices. \cite{tang2025ultrapangu} proposes a novel architecture that intrinsically achieves load balancing across devices. By partitioning experts into device-specific groups and enforcing a routing strategy that activates a fixed number of experts per group, the architectural design ensures balanced computational workloads.
    \item \textbf{Expert Capacity:} Although auxiliary loss provides a gradient-based soft method for load balancing, the physical VRAM constraints necessitate enforced load balancing mechanism. Therefore, expert capacity\cite{lepikhin2020gshard,fedus2022switch} is introduced. Expert capacity refers to the threshold of how many tokens an expert can process. It is calculated by evenly distributing the number of tokens in a batch across the number of experts, and then multiplying by a capacity factor. A capacity factor greater than 1.0 creates additional buffer to accommodate for when tokens are not perfectly balanced across experts. If too many tokens are routed to an expert, the excess tokens are considered overflow, have their computation skipped, and are sent to the next layer via residual connections or completely discarded. \cite{fedus2022switch} defines expert capacity as:
\begin{equation}
expert \ capacity = \left ({ \frac{T}{E}} \right) { × f}
\end{equation}
Where \( {T} \) is the number of tokens, \( {E} \) is the total number of experts, and \( {f} \) is a free hyperparameter called the capacity factor. As shown in \fig{fig:dynamicrouting}, the capacity factor f introduces a trade-off: If the capacity factor is too large, this wastes computational resources through excessive padding. Conversely, if the capacity factor is too small, this sacrifices model performance due to token discards.\\
\begin{figure}[H]
    \centering
    \includegraphics[width=1.0\linewidth]{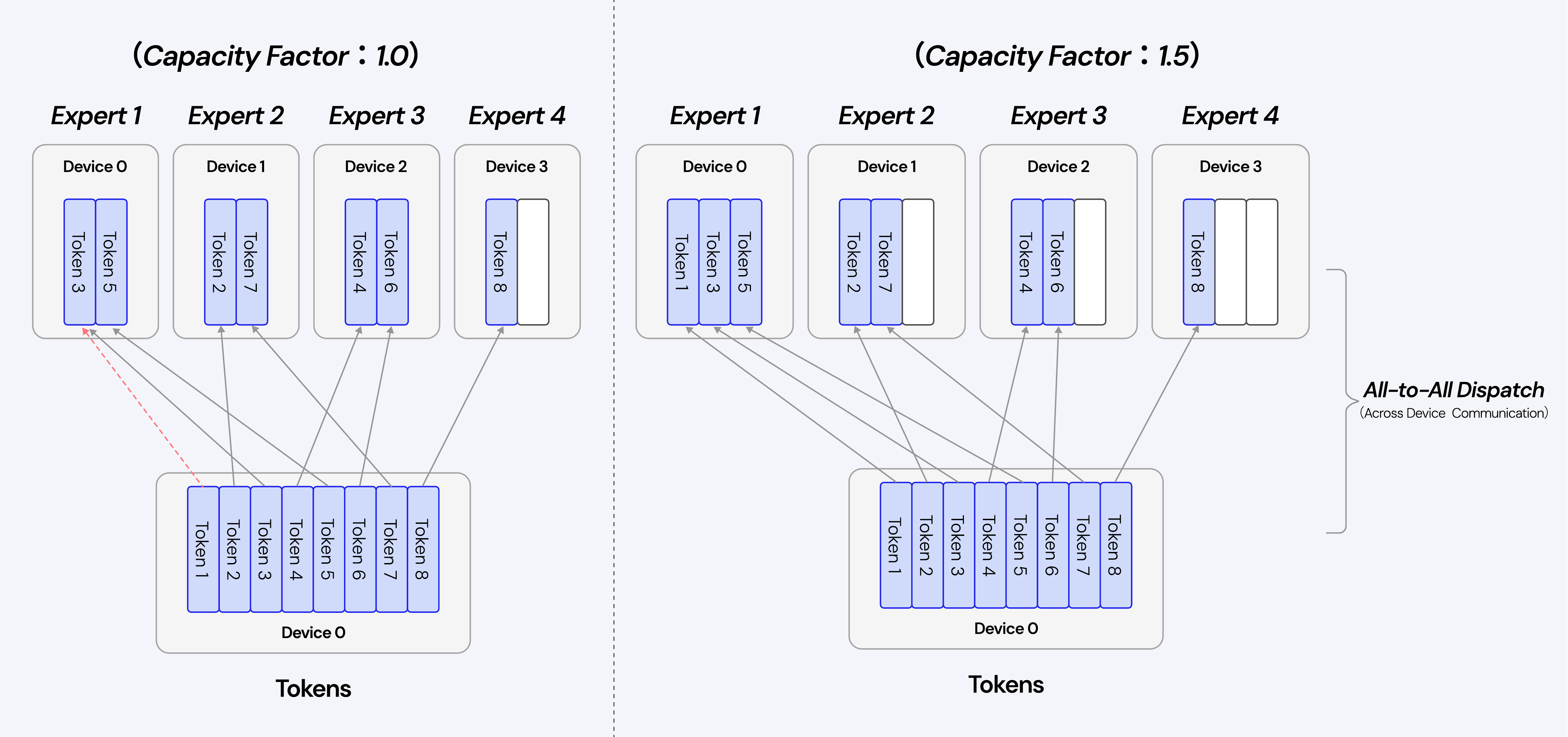}
    \caption{Illustration of token routing dynamics. The tokens are split across all devices using data parallelism. The experts are split across different devices, and each expert processes a fixed expert capacity of tokens modulated by the capacity factor. For a clear and intuitive view, the figure only shows the All-to-All Dispatch of Device 0. With \( {T} \)=8, \( {E} \)=4, it allows the router to send up to 2 tokens per expert. When the capacity factor \( {f} \) is 1.0, it must discard one token (red arrow). When the capacity factor \( {f} \) is 1.5, it needs to add excessive padding (white rectangles/empty slots).}
    \label{fig:dynamicrouting}
\end{figure}     
\end{itemize} 

\subsection{Expert Network} \label{sect:s2dot2}
The expert network is a core component of the MoE architecture. By splitting the dense FFN layers into multiple independent parts, each of which is an independent neural network—expert network\cite{shazeer2017outrageously}. Each expert possesses specialized knowledge in distinct domains and handles specific computational tasks and data subsets. These expert networks collectively accomplish a given task by each processing dedicated partitions of the input data. In practice, these experts are typically FFN units, but they can also be CNN\cite{wang2020deep,lou2021sparse} or more complex network structures.

\subsubsection{Shared Expert} \label{sect:s2dot2dot1}
\begin{figure}[H]
    \centering
    \includegraphics[width=1.0\linewidth]{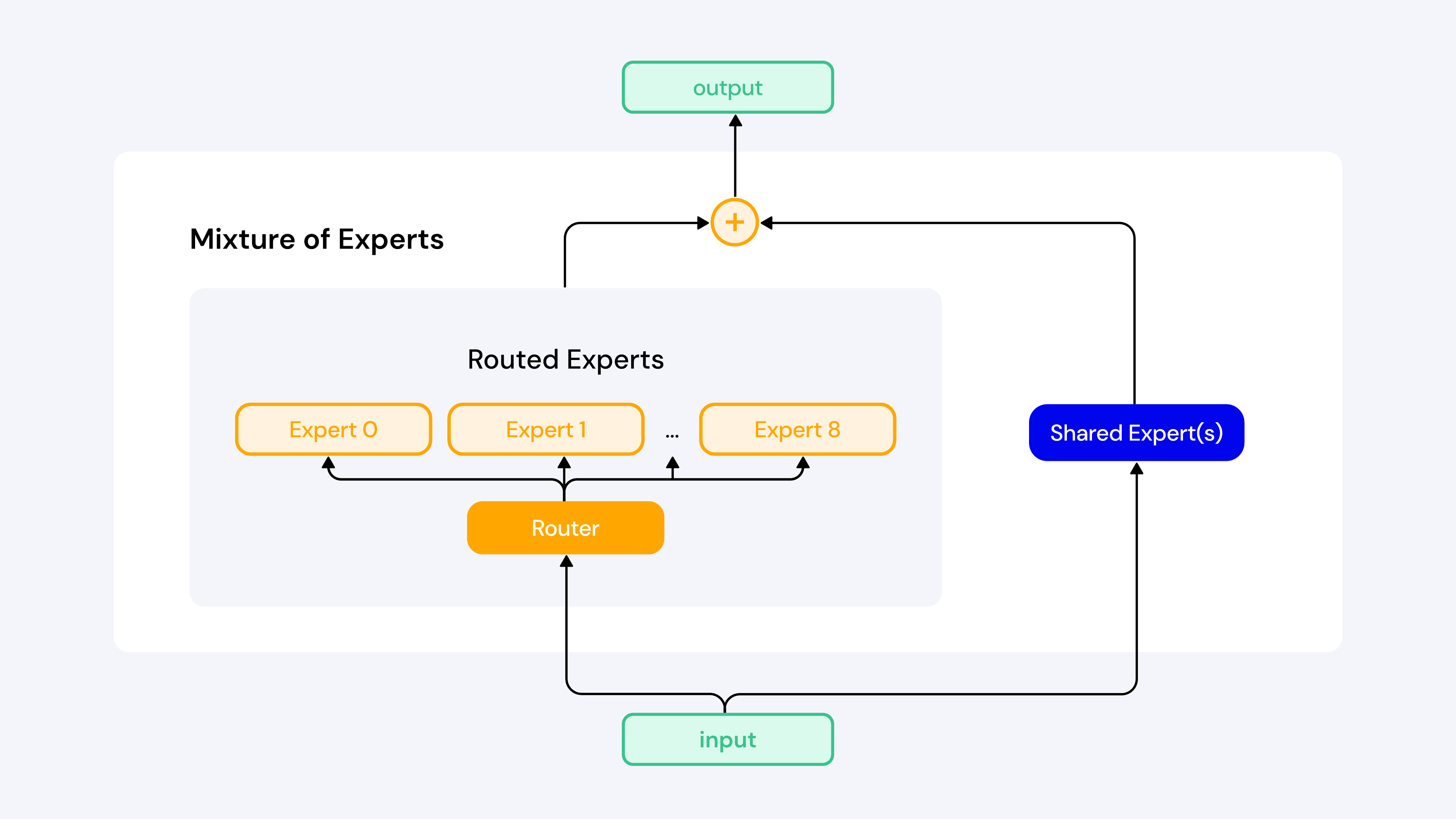}
    \caption{Expert network architecture with shared expert.}
    \label{fig:sharedexpert}
\end{figure}

Unlike conventional MoE, where all experts are designed independently, \seqsplit{DeepSeekMoE}\cite{dai2024deepseekmoe} introduces a fine-grained experts' method. The model further divides the expert network into shared experts and router experts. Shared experts are responsible for processing general features of all tokens, while routing experts dynamically allocate tokens based on their specific characteristics. This division not only reduces model redundancy and improves computational efficiency, but also ensures that shared experts capture common knowledge. The number of shared experts is fixed and always active, as shown in \fig{fig:sharedexpert}.\\

\indent In conventional routing strategies, tokens assigned to different experts may require shared knowledge or information. As a result, multiple experts might converge to learn the same shared knowledge in their parameters, leading to parameter redundancy. However, if there are dedicated experts responsible for capturing and sharing knowledge, this redundancy across different contexts can be alleviated. This reduction in redundancy helps build more parameter-efficient models\cite{qwen_moe,llamamoe,qiu2025demonsdetailimplementingload,tang2025pangu,KimiK2}.

\subsubsection{Interpretability} \label{sect:s2dot2dot2}
\indent MoE models more naturally lend themselves to interpretability studies\cite{akrour2021continuous,mustafa2022multimodal,pavlitska2023sparsely}, because each input is processed by an identifiable, discrete subset of the model weights. For example, some experts may specialize in handling punctuation marks, while others focus on proper nouns or similar elements. Additionally, researchers conducted multilingual training on the model\cite{zoph2022st,mixtureofexperts}.

\indent Interpretability of MoE models has not only been limited to text. \cite{mustafa2022multimodal} is a multi-modal model that was observed to learn experts that specialize in textual and visual data, including patches of textures, plants, eyes, and words.

\section{Centralized and Decentralized Paradigm} \label{sect:s3}
As the remarkable capabilities of large models have elevated artificial intelligence technology to a new level, the AI industry has entered a new round of intense competition, which is becoming increasingly fierce. In the current AI field, the scale of training data and model parameters is showing a growing trend. The latest generation of MoE models that have larger numbers of parameters and longer context windows, which enables them to perform more complex cognitive tasks across a larger knowledge base.\\
\indent Some of the most recent models\cite{yang2025qwen3,liu2024deepseek,KimiK2}, however, have scaled to 1 trillion parameters, have context windows that exceed 128K tokens, and have multiple feedforward networks (experts) that can operate independently. These models cannot fit on a single GPU, which means that the models must be chopped into smaller chunks and parallelized across multiple GPUs.\\
\indent Traditional training and inference of MoE models are distributed on high-performance dedicated computing clusters via high-speed RDMA network. Most of current advanced and high-efficient frameworks \cite{shoeybi2019megatron,li2023colossal,rajbhandari2022deepspeed,10.1145/3600006.3613165,deepep2025} focus on training and inference in homogeneous data center environments, are referred to as centralized paradigm. However, for resource-limited of individual researchers and small laboratories, even for well-resourced organizations, the prohibitive costs of high-performance dedicated computing clusters are unaffordable. Only a handful of large research and development corporations and institutions possess sufficient computational resources to develop advanced MoE models, leading to a quasi-monopoly that hindering people who lack large-scale high-end GPUs from training or deploying MoE models. In fact, consumer-level GPUs, which constitute a larger market share, are typically overlooked due to their weaker computing performance and lower communication bandwidth. Additionally, users may have privacy concerns. Decentralized paradigm has emerged as a promising paradigm to leverage dispersed resources across individual consumer-level GPUs and clusters, offering the potential to democratize MoE models development for broader communities. Collaborative computing allow for a much faster, smarter and scalable way of handling complex tasks. In this section, beyond centralized paradigm, we will delve into decentralized paradigm unlocking the potential collaborative computing of training and inference of MoE models with privacy protection in heterogeneous environments, while delineating critical challenges faced by the decentralized paradigm and reviewing existing related research efforts addressing these issues. Notably, efficient centralized paradigm methods remain essential in the decentralized paradigm, where resource constraints pose greater challenges.

\subsection{Centralized Paradigm} \label{sect:s3dot1}
Currently, MoE models training and inference primarily employ centralized resources deployed on high-performance dedicated computing clusters interconnected via high-speed RDMA networks. Nevertheless, efficient resource utilization remains challenging even in dedicated centralized infrastructures. In this section, we will comprehensively review the state-of-the-art methods for optimizing resource utilization. \tabref{tabref:moe_censystem} presents an overview of highly efficient centralized frameworks that employ advanced resource optimization methods.

\begin{table}[H]
\centering
\caption{An overview of the highly efficient centralized frameworks.}
\resizebox{1\textwidth}{!}{
\renewcommand\arraystretch{1.2}
\begin{tabular}{l|cccc|c} 
\toprule
{\textbf{Papers}} & \multicolumn{4}{c|}{\textbf{Hybrid Parallelization}} & {\textbf{Time}} \\ 
 \cline{2-5} & Expert Parallelism & Data Parallelism & Pipeline Parallelism & Tersor Parallelism & \\ 
\midrule
Switch \cite{fedus2022switch} & \checkmark &	\checkmark & & \checkmark & 2022.01 \\
FasterMoE \cite{he2022fastermoe} & \checkmark &	\checkmark & \checkmark & & 2022.03 \\
Megablocks \cite{gale2023megablocks}  & \checkmark & & & \checkmark & 2023.05 \\
Tutel \cite{hwang2023tutel} & \checkmark & \checkmark & & \checkmark & 2023.05 \\ 
DeepSpeed-TED \cite{singh2023hybrid} & \checkmark & \checkmark & &  \checkmark & 2023.06 \\
SmartMoE \cite{zhai2023smartmoe}  & \checkmark & \checkmark & \checkmark & \checkmark & 2023.07 \\
vLLM \cite{10.1145/3600006.3613165}  & \checkmark & \checkmark & \checkmark & \checkmark & 2023.10 \\
ColossalAI-MoE(OpenMoE) \cite{xue2024openmoe}  & \checkmark & \checkmark & & \checkmark & 2024.02 \\ 
Skywork-MoE \cite{wei2024skyworkmoedeepdivetraining} & \checkmark & & & \checkmark & 2024.06 \\ 
DeepEP \cite{deepep2025}  &	\checkmark & & & & 2025.02 \\ 
\bottomrule
\end{tabular}
}
\label{tabref:moe_censystem}
\end{table}

\subsubsection{Expert Parallelism} \label{sect:s3dot1dot1}
\begin{figure}[H]
    \centering
    \includegraphics[width=1.0\linewidth]{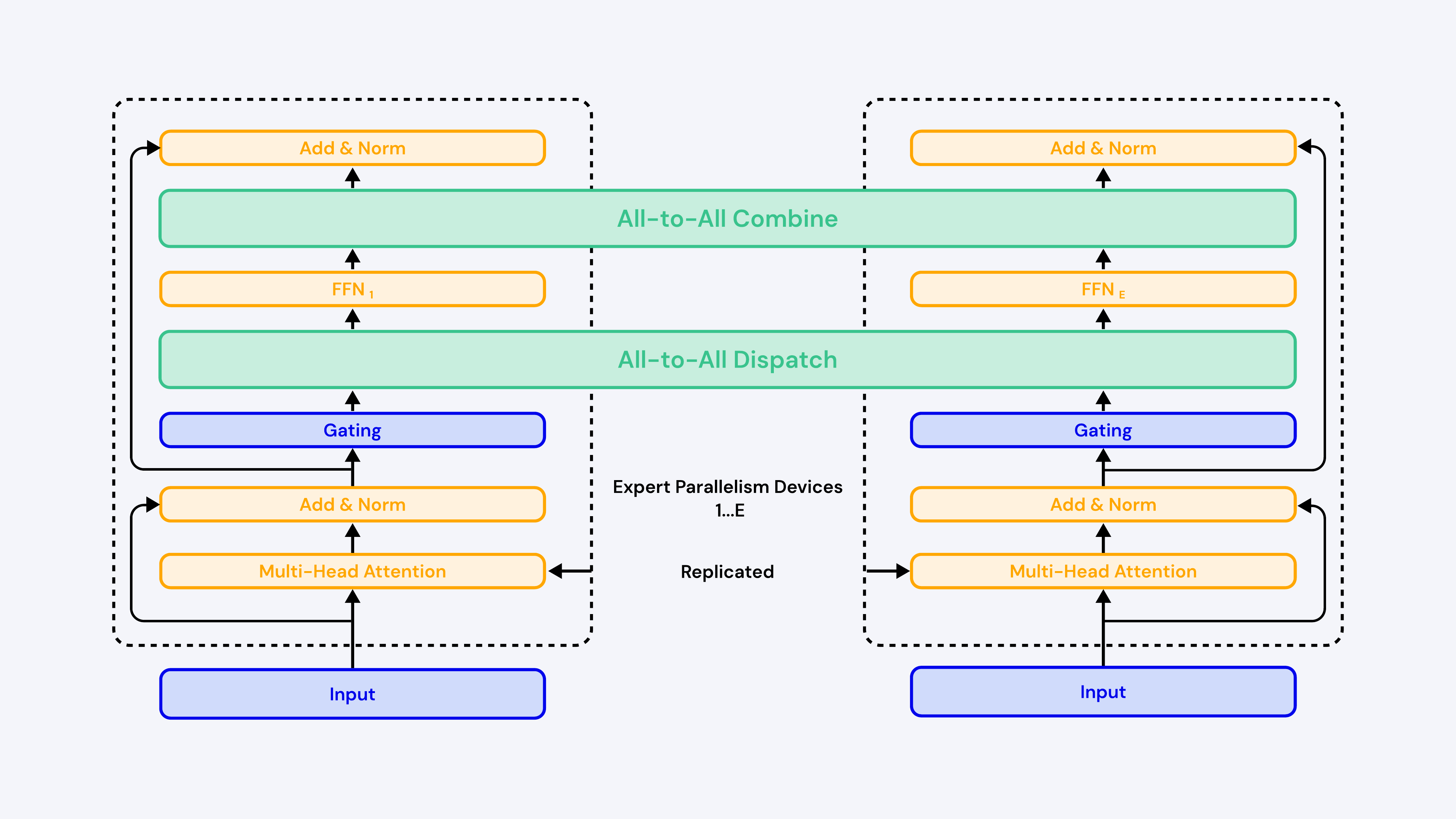}
    \caption{Expert parallelism for the MoE transformer block.}
    \label{fig:expertparallelism}
\end{figure}

Distributed training becomes a must to train MoE models, as the model is so large that it cannot be held in the memory of any single device. To support the distributed training, GShard\cite{lepikhin2020gshard} designs a specific method of parallelism for MoE models, namely expert parallelism (EP). EP assigns different experts to distinct computing devices. Each device sends its own data to the device where the desired experts reside based on the MoE model's routing rules. This greatly reduces the number of parameters that each computation must interact with, as some experts are skipped. For non-MoE layers, EP behaves the same as common parallelism. As shown in \fig{fig:expertparallelism}, the model is split up across the dimension of the experts’ indices, and the input and output features are split along sample dimension. During EP, all-to-all communication is performed to dispatch the input samples to the desired expert models and put the output back to its original location, however, as the scale and frequency of all-to-all communication grow exponentially, the communication time increases significantly, resulting in reduced overall training efficiency. \cite{deepep2025} develops a communication library tailored for MoE and EP to alleviate all-to-all communication bottlenecks, particularly among GPUs. \cite{wang2025toward} introduces Asynchronous Expert Parallelism (AEP), a new paradigm that decouples layer execution from barrier-style synchronization, effectively addresses the GPU underutilization and synchronization bottlenecks that commonly arise in expert parallel MoE serving.

\subsubsection{Hybrid Parallelization} \label{sect:s3dot1dot2}
To further improve the performance of expert parallelism, Hybrid Parallelism has been extensively studied\cite{kim2021scalable,he2022fastermoe,fedus2022switch,zhai2023smartmoe,singh2023hybrid,gale2023megablocks,wei2024skyworkmoedeepdivetraining,xue2024openmoe}, which combines a few of the below parallel strategies to better fit specific models and particular training hardware. \\
\indent Listed below are three common ways of parallelism.

\begin{itemize}
    \item \textbf{Data Parallelism(DP):} The data parallelism (DP) method\cite{li2020pytorch,rajbhandari2020zero} hosts multiple copies of the MoE model on different GPUs or GPU clusters. Forward and backward computation are completed independently on each GPU. Gradients on different GPUs are aggregated before being used in the optimization of the model. However, DP alone is usually not sufficient with the latest generations of MoE modesl, as their model weights don’t fit on a single GPU memory.
    \item \textbf{Pipeline Parallelism(PP):} The model is divided into multiple stages and executes them sequentially on different devices \cite{huang2019gpipe,narayanan2019pipedream}. Each devices stores the parameters of its corresponding stage. The first device reads batches of the data, and devices with adjacent stages exchange intermediate results for forward or gradients for backward computation. To be efficient, PP has to improve the pipeline scheduling of devices to reduce idle waiting time, known as bubble time—a problem that has been intensively studied by prior works\cite{guan2019xpipe,qi2024zero,narayanan2021efficient,arfeen2024pipefill,guan2025pipeoptim}. However, the layer-wise dependency of forward and backward processes limits the scalability of PP\cite{guan2024advances}.
    \item \textbf{Tensor Parallelism(TP):} It vertically splits model stages across multiple devices, each device stores a part of the parameters of the operators and conducts part of its computation, e.g. a tile of a matrix. TP of different operators needs to be designed specifically by experts, and the partitioning method is critical to distributed training performance. Megatron\cite{shoeybi2019megatron} provides the best practice of 1D TP on transformer models. However, 1D TP imposes higher demands on communication speed. Subsequent studies\cite{xu2023efficient,wang20212,bian2021maximizing} have sought to refine this approach for most efficient, and these enhanced tensor parallelism algorithms demonstrate seamless compatibility with pipeline parallelism methods. For Transformer-based models, TP can enhance processing capabilities by allocating more GPU resources, speeding up processing time.    
\end{itemize}

\subsection{Decentralized Paradigm} \label{sect:s3dot2}
\begin{figure}[H]
    \centering
    \includegraphics[width=1.0\linewidth]{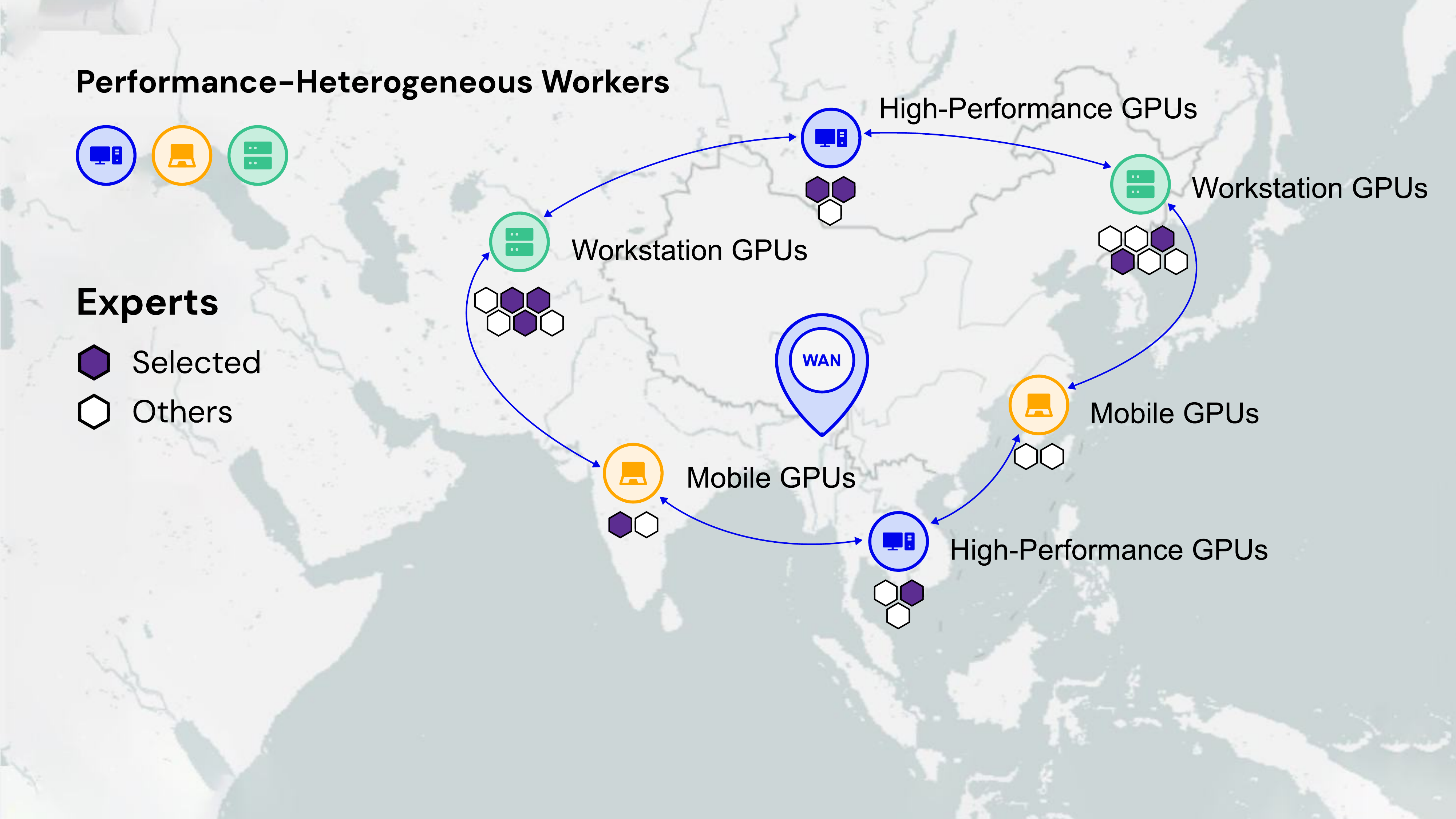}
    \caption{High-level scheme of decentralized MoE.}
    \label{fig:dmoed}
\end{figure}

Decentralized paradigm leverages a broader range of resources compared to traditional centralized paradigm, yet it still employ mentioned above similar parallel strategies. The traditional centralized paradigm, a MoE system defines a machine learning model comprising multiple “experts”, where each expert specializes in solving individual tasks. In other words, it splits complex models or tasks among smaller, more specialized networks. Each expert is trained on a specific aspect of the bigger task or data subset. A decentralized mixture of experts (DMoE) system takes it a step further. Instead of one central “boss” deciding which expert to use, multiple smaller systems each make their own decisions. This means the system can handle tasks more efficiently across different parts of a large system. The decentralized computing architecture enables the DMoE model to operate across laptops, workstations, and data centers by letting each part work independently, thus making it faster and more scalable, as depicted in \fig{fig:dmoed}.\\ 

\indent However, compared with training and deploying MoE models in data centers equipped with high-ends GPUs, decentralized paradigm encounters several critical challenges, encompassing hardware performance heterogeneity, limited communication bandwidth, fault tolerance, security and privacy protection. In the subsequent discussion, we illustrate these challenges, concurrently reviewing existing related research efforts addressing these issues, as summarized in \tabref{tabref:paradigm}.
\begin{table}[H]
    \centering
    \caption{An overview of critical challenges in the decentralized paradigm.}
    \resizebox{\textwidth}{!}{
    \begin{tabular}{ccc}
     \toprule     
     \textbf{Challenges} & \textbf{Works} & \textbf{Key features}  \\  
     \midrule
     \textbf{Performance Heterogeneity} & \makecell[c]{DMoE\cite{ryabinin2020towards}, DeDLOC\cite{diskin2021distributed},Petals\cite{borzunov2022petals}, FusionAI\cite{tang2023fusionai},\\
     ATOM\cite{wu2024atom}, Ravnest\cite{menon2024ravnest}, HeterMoE\cite{wu2025hetermoe}} & \makecell[c]{Ranging from mobile GPUs to \\mid-end and high-performance GPUs.}\\
     \midrule
     \textbf{Communication Efficiency} & \makecell[c]{Saurabh Agarwal et al.\cite{agarwal2022utility}, Moshpit\cite{ryabinin2021moshpit}, SWARM\cite{ryabinin2023swarm},\\
     Diloco\cite{douillard2023diloco}, Opendiloc\cite{jaghouar2024opendiloco}, Dilocox\cite{qi2025dilocox},  StellaTrain\cite{lim2024accelerating}} & \makecell[c]{Instead of high-speed RDMA network,\\compute nodes communicate mainly \\through LANs or even WANs \\with bandwidths under 10 Gb/s.}\\
     \midrule
     \textbf{Fault Tolerance} & \makecell[c]{Kademlia\cite{maymounkov2002kademlia}} & \makecell[c]{A compute node can join at any time, \\while existing nodes may fail to \\process a task for a variety of reasons.}\\
     \midrule
     \textbf{Security and Privacy} & \makecell[c]{Zhizhi Peng et al.\cite{peng2025survey}, ZK-LLMs\cite{wellington2024basedai}, PC-MoE\cite{zhang2025memory}, \\
     Hongyang Zhang et al.\cite{zhang2024towards}, Zhibo Xing et al.\cite{xing2025zero}} & \makecell[c]{Vulnerable to attacks from \\malicious compute nodes that can \\disrupt system operations and \\compromise model robustness.}\\       
     \bottomrule
    \end{tabular}
    }  
    \label{tabref:paradigm}
\end{table}

\subsubsection{Hardware Performance Heterogeneity} \label{sect:s3dot2dot1}
In decentralized paradigms, when aggregating resources across consumer-grade GPUs, multiple nodes, or clusters, these computational devices are typically configured with various memory sizes, and communication bandwidths, ranging from mobile GPUs to mid-end and high-performance GPUs, these resources exhibit high heterogeneity. To prevent slower ones from becoming bottlenecks significantly, \cite{ryabinin2020towards} proposes Decentralized Mixture of Experts(dMoE)-a layer designed for training with vast amounts of unreliable consumer-grade devices, different consumer-grade devices host varying numbers or scales of experts according to its available memory and compute capacity. \cite{diskin2021distributed} designs a novel algorithm framework named DeDLOC which combines parameter servers\cite{10.5555/2685048.2685095}, All-Reduce SGD\cite{sergeev2018horovod}, decentralized SGD\cite{lian2017can} and BytePS\cite{jiang2020unified}. DeDLOC can accommodate a large number of heterogeneous devices with uneven compute capabilities for collaborative training. To address hardware heterogeneity challenges in decentralized training, Petals\cite{borzunov2022petals} proposes server dynamically select responsible layers based on available GPU memory. This is coupled with 8-bit quantization and load balancing to mitigate discrepancies in memory and computational resources on heterogeneous devices. \cite{tang2023fusionai} dissects directed acyclic graphs (DAGs) of model execution into sub-DAGs and loading them onto devices with limited memory. \cite{wu2024atom} presents ATOM, a resilient distributed training framework designed for asynchronous training of vast models in a decentralized setting using cost-effective
hardware, including consumer-grade GPUs and Ethernet. The motivation behind ATOM's design is that a complete model can be executed on a single GPU layer by layer via memory swapping. By profiling individual model layers in detail, it devises an optimal model swapping schedule that effectively addresses the swapping overhead issue. \cite{menon2024ravnest} directly tackles bottlenecks arising from heterogeneous device performance during decentralized training. It proposes a "cluster-first, parallelize-later" strategy that employs genetic algorithm-based clustering to group heterogeneous nodes into capability-similar clusters according to RAM and bandwidth metrics, implements bubble-free asynchronous parallelism within clusters to eliminate straggler-induced idleness through asynchronous model execution, and facilitates efficient cross-cluster synchronization via multi-ring All-Reduce communication with parallelized ring coordination. \cite{wu2025hetermoe} presents HeterMoE, a system to efficiently train MoE models on heterogeneous GPUs with no extra communication. HeterMoE disaggregates attention and expert modules to fully utilize each GPU’s capability. HeterMoE introduces zebra parallelism (ZP), along with asymmetric expert assignment (Asym-EA), to enable computation overlapping and fine-grained load balancing.\\
\subsubsection{Communication Efficiency} \label{sect:s3dot2dot2}
Remote Direct Memory Access(RDMA) technology has found extensive deployment in modern data centers, offering low latency and high throughput benefits that are particularly advantageous for generative AI training. Typically, centralized training is carried out in datacenter-grade GPU clusters equipped with homogeneous high-speed RDMA network, and network can reach over 400 Gb/s. In contrast, Decentralized training communicate through LANs or even WANs with bandwidths under 10 Gb/s. This will make training extremely slow, as gradient exchange is often bottlenecked by scarce network bandwidth, eventually leading to GPU underutilization\cite{agarwal2022utility}. Such bottlenecks are even worse in hybrid cluster settings when train a single model collaboratively across multiple geographically distributed consumer-grade GPUs, cloud GPU instances separated by the WANs with constrained and highly variable bandwidth.\\ 
\indent Due to bandwidth limitations of LANs and WANs, it is essential to optimize communication efficiency for decentralized training to alleviate communication bottlenecks. \cite{ryabinin2021moshpit} proposes a "Dynamic Grouping-AllReduce" mechanism that dynamically group active nodes into small-scale clusters. Within each cluster, it performs bandwidth-optimal AllReduce for gradient averaging, then iteratively propagates results across clusters to alleviate bandwidth constraints in decentralized environments. \cite{ryabinin2023swarm} proposes the SWARM parallel algorithm, the first decentralized model parallelism approach that leverages stochastic fault-tolerant pipelining and dynamic rebalancing to enhance communication efficiency. This approach deploys multiple candidate devices per pipeline-parallel stage. When a device outperforms others, it concurrently processes inputs from multiple slower predecessors and distributes outputs to multiple slower successors, maximizing bandwidth utilization.
\cite{douillard2023diloco,jaghouar2024opendiloco} introduce two key innovations to integrate decentralized computing resources into a virtual supercomputer. One is low-frequency synchronization that global synchronization occurs only every H steps, maintaining parameter stability via momentum updates, and the other is decoupled inner/outer optimization that local replicas independently perform H-step internal optimization (e.g., AdamW), then aggregate updates through an external optimizer (e.g., Nesterov momentum SGD). This simultaneously reduces communication frequency and enhances tolerance to large batch sizes. \cite{qi2025dilocox} enables low-communication decentralized cluster training for up to 100-billion-parameter models by integrating gradient compression with overlapped communication and local training.\\
\indent \cite{lim2024accelerating} presents StellaTrain, the first framework for distributed training that minimizes the communication intensity of model training in multi-cluster environments separated by a WAN. StellaTrain introduces two key enablers to achieve such high training speeds. First, StellaTrain employs gradient compression to effectively use the network in low-bandwidth environments and exploits the resulting sparsity of gradients to devise computationally efficient compression and optimization. Second, a layer-wise partial staleness mechanism is designed in StellaTrain, where some layers receive gradient updates immediately, while others are delayed by one iteration. 

\subsubsection{Fault Tolerance} \label{sect:s3dot2dot3}
With distributed decision-making, the system can continue functioning even if one gate or expert fails. This resilience ensures uninterrupted operation and minimizes the risk of complete system failure. Given the inherent instability of computational resources contributed by communities, node can join at any time, while existing nodes can exit due to various reasons. Moreover, node failures and communication disruptions are inevitable in decentralized infrastructure. Consequently, fault tolerance becomes a critical requirement to ensure stable and efficient training processes.\\
\indent To implement fault tolerance mechanisms in a decentralized system, distributed hash table (DHT) have emerged as a core technology. This is a family of distributed data structures that store key-value pairs across multiple computers in a network. A single computer within such structure only needs to "know" O(log N) out of N computers, at the same time it can look up any key with at most O(log N) requests to his peers. There are several DHT variants, but they all have common properties:
\begin{itemize}
    \item \textbf{Decentralization:} Nodes form and maintain DHT without any central coordination.
    \item \textbf{Scalability:} DHT can scale to millions of active nodes that are continually joining and leaving.
    \item \textbf{Fault tolerance:} A failure in one or a few nodes does not affect DHT integrity and availability.    
\end{itemize}

\indent By far, the most popular DHT variation is Kademlia\cite{maymounkov2002kademlia} with numerous applications such as BitTorrent, I2P, and Ethereum.

\subsubsection{Security and Privacy Protection} \label{sect:s3dot2dot4}
With the rapid advancement of generative AI technologies, data privacy and model security in decentralized environment have become critical challenges. The potential risks of infringement privacy are the unauthorized data access and manipulation. There is also the threat of malicious computing nodes that can disrupt system operations and compromise model robustness. The issue of membership inference can expose if a specific data point was used in training the model. In response to these challenges, privacy-preserving techniques such as secure multiparty computation, differential privacy, and encryption are suggested as mitigation strategies.\\
\indent Zero-Knowledge Machine Learning (ZKML) is an emerging technology that applies Zero-Knowledge Proofs (ZKP) to the machine learning domain\cite{peng2025survey}. It enables data owners to utilize their data for training machine learning models without sharing raw data with third parties. This approach ensures data privacy and mitigates the risk of data breaches. Simultaneously, it allows data owners to selectively share model outcomes, thereby reconciling data security requirements with machine learning operational needs. It offers a viable solution for enhancing security and privacy protection in decentralized environment\cite{wellington2024basedai,zhang2024towards,xing2025zero}.\\
\indent \cite{zhang2025memory} proposes Privacy-preserving Collaborative Mixture-of-Experts (PC-MoE), which enables multiple parties to leverage each other’s compute resources and data to reduce the hardware burden and improve their own MoE model’s performance without sacrificing data privacy. It demonstrates that preserving privacy does not always have to come at a steep cost to utility or vice versa: by routing only sparse expert signals, parties obtain near-centralized performance, enjoy lower hardware requirements, while revealing virtually nothing to an attacker.\\
\indent These findings motivate future privacy research to explore how far the utility frontier can be pushed with minimal compromise in privacy and safety.
\section{DownStream Applications} \label{sect:s4}
\begin{figure}[H]
    \centering
    \includegraphics[width=1.0\linewidth]{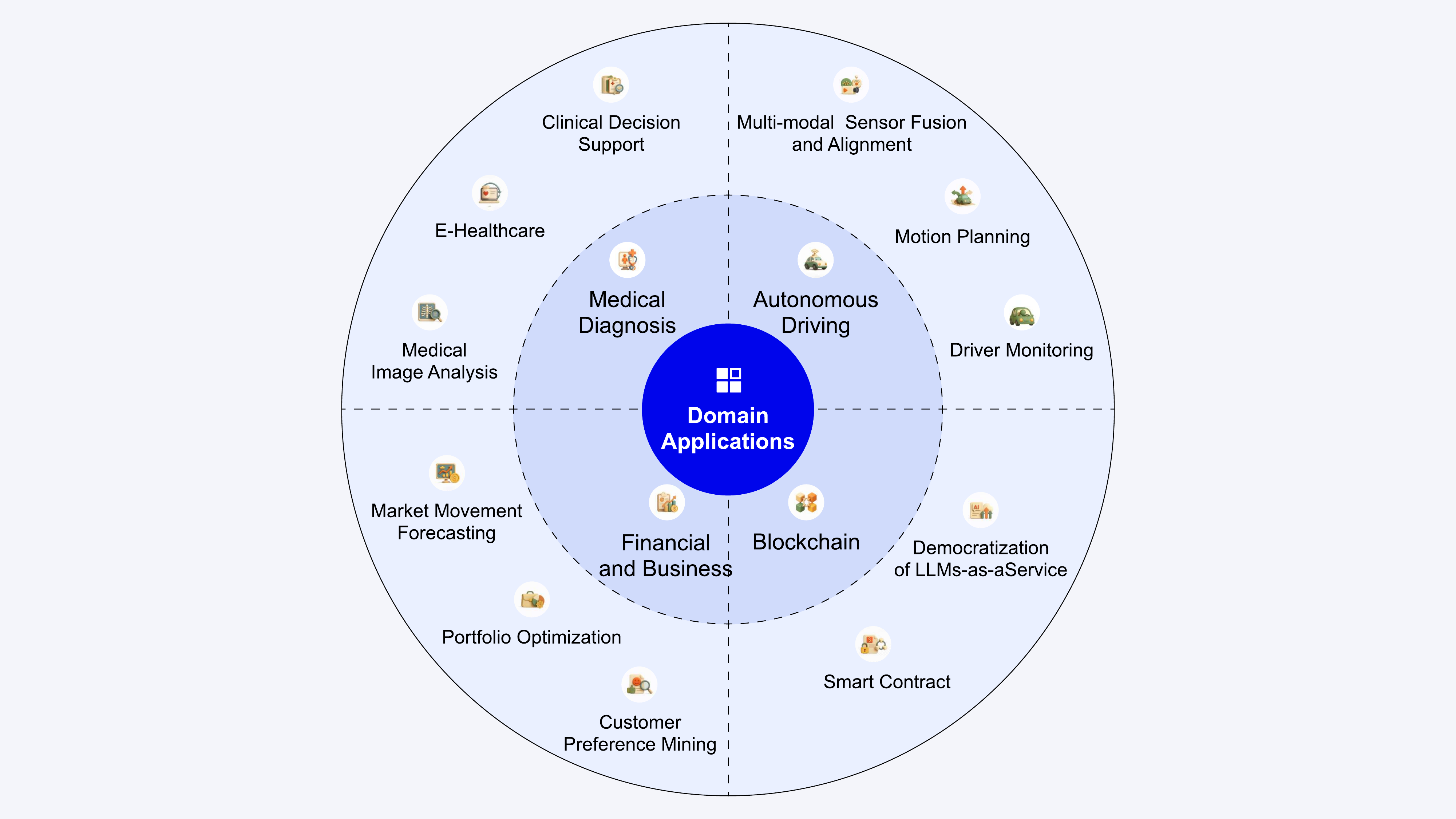}
    \caption{An overview of MoE applications in typical vertical domains.}
    \label{fig:applications}
\end{figure}

In recent years, MoE technology has witnessed exponential advancement, with deployments now spanning NLP\cite{llamamoe,DBRX}, CV\cite{riquelme2021scaling,zhang2024clip}, multimodal systems \cite{mustafa2022multimodal,lin2024moe,wu2024deepseek,han2024vimoe,li_unimoe}, and intelligent agent frameworks, leveraging its ability to dynamically allocate computational resources and specialize in different data distributions, demonstrating revolutionary efficiency advantages\cite{liu2024deepseek}. In the following, we will focus on exploring the typical vertical domains applications of MoE, including medical-assisted diagnosis, autonomous driving, financial analysis, business intelligence, and blockchain, as illustrated in \fig{fig:applications}. The aim is to provide an overall understanding of how MoE can be utilized for specific tasks. These application scenarios are also more appropriate for decentralized infrastructures, where resource constraints are more severe and complex.
\subsection{Medical Diagnosis}  \label{sect:s4dot1}
Recent advancements in general-purpose or domain-specific multimodal models have witnessed remarkable progress for medical decision-making. However, medical data is inherently heterogeneous across different resolutions, modalities and clinical centers\cite{huo2021sparse}, posing unique challenges for developing generalizable foundation models. Conventional entails training distinct models per dataset or using a shared encoder with modality-specific decoders, but these approaches incur heavy computational overheads and suffer from poor scalability, hindering their clinical utility across diverse resource-constrained scenarios in practice.\\
\indent \cite{jiang2024med} proposes a lightweight framework for multimodal medical tasks, addressing both discriminative and generative needs. Optimized for resource-constrained environments, it involves aligning medical images with language model tokens, task-specific instruction tuning, and domain-specific expert fine-tuning. From coarse global patterns to fine-grained localized structures processing framework is introduced in \cite{chopra2025medmoe}, a vision–language model that dynamically routes multi-scale visual features through a diagnosis-conditioned MoE framework. This framework produces localized visual representations aligned with textual descriptions, without requiring modality-specific supervision at inference, improves the performance of modality-specialized visual representations in clinical vision-language systems. \cite{liu2024medical-e} introduces a novel approach to improve medical text embeddings tasks for Retrieval-Augmented Generation (RAG) in specialized medicine domains. Med-MoE-Embed leverages a pretrained embedding backbone augmented with a trainable MoE network, allowing for efficient adaptation to specific medical subdomains and tasks,significant improvements in retrieval accuracy and generation quality, enhancing the model’s adaptability across various medical datasets. Med-MoE-Embed mitigates the challenges of limited data accessibility and domain-specific requirements in the medical field, offering a versatile and efficient solution for enhancing embedding quality in medical natural language processing applications. \cite{xie2025dflmoe} proposes a decentralized federated learning framework named dFLMoE for medical data analysis. In this framework, clients directly exchange lightweight head models with each other. After exchanging, each client treats both local and received head models as individual experts, and utilizes a client-specific MoE approach to make collective decisions. This design not only protects patients' privacy but also removes the dependency on the central server to enhance the robustness of the framework.\\
\indent \cite{huo2021sparse} introduces MoE and specifically couple it with a sparse gating network model to handle patient heterogeneous electronic health record (EHR) for risk prediction.it enhance the effectiveness of risk prediction.\cite{yuan2024efficient} addresses the difficulties of MoE models deploy on the Internet of Medical Things (IoMT) for individuals’ personalized e-healthcare. By designing a new medical LLM based on the MoE architecture with an offloading strategy, meeting the deployment resource demands in the IoMT, improving e-healthcare services and the privacy protection for users.\\
\indent \cite{jiang2024m4oe,liu2024medical,chen2024low} focus on medical image segmentation base on MoE. Notably,\cite{jiang2024m4oe} proposes the Medical Multimodal Mixture of Experts (M4oE) framework named M4oE, leveraging the SwinUNet architecture. Specifically, M4oE comprises modality-specific experts; each separately initialized to learn features encoding domain knowledge. Subsequently, a gating network is integrated during fine-tuning to modulate each expert’s contribution to the collective predictions dynamically. This enhances model interpretability and generalization ability while retaining expertise specialization. Simultaneously, the M4oE architecture amplifies the model’s parallel processing capabilities, and it also ensures the model’s adaptation to new modalities with ease. Moreover, M4oE showcases a significant reduction in training overhead.\\
\subsection{Autonomous Driving} \label{sect:s4dot2}
The great success of sparsely-gated MoE in various fields has inspired their application in autonomous driving. End-to-end learning from sensory data has shown promising results in autonomous driving. While employing many sensors enhances world perception and should lead to more robust and reliable behavior of autonomous vehicles, it is challenging to train and deploy such network and at least two problems are encountered in the considered setting. The first one is the heterogeneous data typically originates from various sensors, systems, exhibiting a high degree of heterogeneity and diversity. The other is the increase of computational complexity with the number of sensing devices.\\
\indent \cite{fang2020multi} proposes a Multi-modal Experts Network, where each sensor is paired with a lightweight expert sub-network, and introduce a two-level gating network to handle inputs coming from three cameras and one LiDAR. The gating network chooses the camera input in a discrete way from among several mutually-exclusive sensors. Alternatively, the network chooses the LiDAR sensor, which covers the same field of view as the camera sensors, and identifies continuously in real-time the part of its depth map with a narrow field of view that is useful for steering autonomously. On resource-constrained in-vehicle platforms, the proposed multi-modal experts network to perform conditional computation that converts multi-sensor redundancy into robustness rather than a computational burden, thereby addressing the fundamental challenges of multi-sensor fusion in autonomous driving systems. \cite{morra2023mixo} presents MIxture of eXperts Odometry (MIXO), a data-driven, machine learning-based technique that loosely combines odometry outputs from multiple cameras to obtain a more accurate and robust global estimate. In MIXO, each camera (expert) is individually processed by a state-of-the-art visual odometry algorithm. Then, the odometry estimates are mixed by a gating network, which selects the locally optimal experts in the current operational conditions and weights their contributions accordingly. MIXO achieves more robust and accurate results than any single camera, reducing the absolute rotational and translation error. \cite{xu2025limoe} introduces a framework that integrates the MoE paradigm into LiDAR data representation learning to synergistically combine multiple representations, such as range images, sparse voxels, and raw points, captures complementary information from different representations, enabling more robust scene understanding.\\
\indent \cite{pini2023safe} is the first work to successfully apply the MoE mechanism to joint prediction-and-planning modeling for safe autonomous driving on public roads. By learning and selecting from a distribution of expert trajectories, it captures multimodal future trajectory distributions, thereby enhancing system safety and robustness in complex traffic scenarios. Compared with traditional approaches, MoE delivers several advantages: strong multimodal modeling capabilities that adapt to intricate traffic situations, a data-driven safety strategy that obviates rule-based systems and is readily scalable, and high interpretability. \cite{feng2025artemis} introduces autoregressive end-to-end trajectory planning with MoE for autonomous driving, it addresses traditional static planners' inability to model temporal dependencies. Through its integrated MoE architecture with dedicated routing networks, ARTEMIS dynamically captures the intrinsic dynamic characteristics of driving behavior and effectively accommodates diverse driving environments. The framework further incorporates a lightweight batch reallocation strategy, significantly enhancing both training efficiency and deployment viability. \cite{sun2024generalizing} applies the MoE architecture to end-to-end motion planning tasks. Through an expert routing mechanism, it dynamically selects specialized expert sub-networks to resolve conflicts and trade-offs between different driving objectives (e.g., obstacle avoidance, yielding, lane keeping), significantly improves generalization across diverse driving scenarios, including challenging out-of-distribution zero-shot cases. This is the first introduction of expert specialization for learning driving reward preferences, effectively resolving "modality collapse" and "reward balancing" conflicts. The resulting scalable motion planning model scales up to 800M parameters, employs 8 experts with top-2 expert activation, and supports seamless scaling from small (100M) to large (1B) models following scaling law. \cite{yang2025drivemoe} introduces both a Scene-Specialized Vision MoE and a Skill-Specialized Action MoE, specifically designed for end-to-end autonomous driving scenarios, named DriveMoE. DriveMoE dynamically selects contextually relevant camera views and activates skill-specific experts for specialized planning.The Vision MoE employs a learned router to dynamically prioritize camera views aligned with the immediate driving context, integrating projector layers that fuse these selected views into a cohesive visual representation. Concurrently, the Action MoE leverages another routing mechanism to engage distinct experts within a flow-matching planning architecture, with each expert dedicated to handling specialized behaviors such as lane following, obstacle avoidance, or aggressive maneuvers. By introducing context-driven dynamic expert selection across both perception and planning modules, DriveMoE significantly enhancing computational efficiency and robustness to rare.\\
\indent \cite{wang2024efficient} designs a innovative multi-task Driver Monitoring System(DMS), termed VDMoE, which leverages expert specialization and task-level gating networks, the model achieves state-of-the-art performance on drowsiness, cognitive load, heart rate, and respiration rate estimation, simultaneously delivering high accuracy, efficiency, and interpretability, thereby providing a reliable monitoring solution for autonomous driving.

\subsection{Financial and Business Analysis} \label{sect:s4dot3}
MoE have revolutionized data analysis domains due to its ability to handle heterogeneous data sources and dynamic changes\cite{wang2024hmoe}. For intricate financial and business analysis tasks, MoE tailored to specific domains can achieve a more comprehensive understanding and more superior insights by routing different factors to specialized expert networks and the gating network dynamically coordinates the outputs of the experts. With the scale and complexity of financial and business data have increased significantly, and traditional analysis tools are struggling to cope with this. Financial and business institutions have gradually adopted the MoE architecture to build an automated data science workflow for improve the accuracy and efficiency of data analysis tasks, helping them effectively process market movement forecasting, risk assessment making, portfolio optimization, fraud detection, customer preference mining, etc. For example, the quantitative investment field is highly dependent on accurate stock forecasting and profitable investment decision-making\cite{sun2022quantitative}. This role-specific specialization enhances the model's ability to integrate their domain-specific expertise.\\
\indent \cite{yu2024finteamexperts} introduced FinTeamExperts, a novel framework of role-specialized LLM designed as a MOE to excel in financial analysis tasks. By mimicking a real-world team setting within the finance domain, each model in FinTeamExperts specializes in one of three critical roles: Macro Analysts, Portfolio Managers, and Quantitative Analysts. The results showcase the potential of advanced LLM in transforming financial analysis and decision-making, forming a comprehensive and robust financial analysis tool, paving the way for more sophisticated and practical AI applications in the finance industry.\\
\indent \cite{vallarino2024dynamic} offers an extensible MoE framework for financial time-series modeling.It applies MoE to stock-price forecasting, combines an RNN for volatile stocks and a linear model for stable stocks, dynamically adjusting the weight of each model through a gating network, significantly improves predictive accuracy across different volatility profiles. \cite{xiao2024autoeis} designs an Multi-field-aware Mixture-of-Experts (MfMoE) architecture which can simultaneously learn the single-field and global-field information, significantly improve the accuracy of default prediction in the financial field, effectively solving the problems of numerical feature encoding and high-order feature interaction modeling.\\
\indent \cite{wei2025deep} introduces a Mixture-of-Experts-based deep reinforcement learning portfolio model (MoEDRLPM) that addresses limitations in spatio temporal modeling and strategy diversity in traditional investment approaches by dynamically selecting the current optimal expert from the mixed expert pool through router. The expert makes decisions and aggregates to derive the portfolio weights. Notably, the model achieves significant improvements in return metrics. These results robustly demonstrate MoE's practical value in financial decision-making scenarios—where expert specialization simultaneously optimizes returns and risk control.\\
\indent Understanding consumer choice is fundamental to marketing and management research, since it allows a business to make better use of marketing budgets as well as to gain a competitive edge over rival companies. More importantly, it demonstrates better knowledge of various customer purchasing behaviors and patterns over time.Some of the challenges faced by e-commerce, stores, and supermarkets involve dealing with huge volumes of customers with different and similar wants, different and similar purchase prices, and buying patterns. \cite{anireh2024model,vallarino2025consumers} focus on customer mining technique based on MoE, as a machine learning-driven alternative that dynamically segments consumers based on latent behavioral patterns. By leveraging probabilistic gating functions and specialized expert networks, MoE provides a flexible, nonparametric approach to modeling heterogeneous preferences, significantly enhances predictive accuracy over traditional econometric models, capturing nonlinear consumer responses to price variations, brand preferences, and product attributes. The findings underscore MoEs potential to improve demand forecasting, optimize targeted marketing strategies, and refine segmentation practices. These studies bridges the gap between data-driven MoE approaches and marketing theory, advocating for the integration of AI techniques in managerial decision-making and strategic consumer insights. As markets continue to evolve and consumer preferences become increasingly complex and dynamic, the MoE framework provides a powerful tool for understanding and predicting economic behavior in an era of data-driven decision-making.

\subsection{Blockchain} \label{sect:s4dot4}
A DMoE model adapts MoE to a decentralized network, such as a blockchain \cite{chong2025llm}. This means that rather than a central entity controlling the experts, the decision-making and control are spread across multiple smaller systems which hosted on peer devices. Simply put, the network autonomously selects the most suitable expert (a node or smart contract) based on what the task needs.\\
\indent Although the MoE model has a long history and achieved great success in many fields during the past few years\cite{jacobs1991adaptive,yuksel2012twenty,eigen2013learning,fedus2022review,jiang2024mixtral}, the intersection of MoE and blockchain is largely under-explored, MoE can still play a role in several aspects of blockchain technology. Blockchain is a innovative decentralized, distributed ledger technology that enables secure and transparent transactions without the need for intermediaries. By utilizing advanced cryptographic methods, blockchain ensures the integrity and verification of each transaction, establishing a highly reliable technological framework. Within this ecosystem, smart contracts function as self-executing programs on the blockchain, automating the management of digital assets such as cryptocurrencies. These contracts are activated when specific conditions are met and, once deployed, become permanent components of the blockchain.\\
\indent \cite{yuan2025mos} proposes a smart contract vulnerability detection framework based on mixture-of-experts tuning (MOE-Tuning) of LLM named MOS. Through the MoE architecture establishes a "Dynamic Routing-Specialized Experts" collaborative framework that reconstructs the paradigm for smart contract security detection, significantly outperforms existing state-of-the-art methods with improvements in accuracy. The vulnerability explanations generated by MOS also demonstrate high quality, achieving positive ratings for correctness, completeness, and conciseness, respectively, through a combined approach of human evaluation and LLM evaluation. It demonstrates the effectiveness and scalability of MoE in smart contract security detection. SAEL\cite{yu2025saelleveraginglargelanguage} further introduce an Adaptive Mixture-of-Experts architecture mechanism which dynamically adjusts feature weights via a Gating Network to enhance the performance of smart contract vulnerability detection.
\section{Challenges and Opportunities} \label{sect:s5}
Although sparse MoE models have undergone continuous exploration and innovation, achieving significant advancements, the main technical challenges still remain.The following are some of the key challenges and promising research directions we have considered.

\begin{itemize}
    \item \textbf{Load balancing:} Load balancing is a critical issue in MoE models training. However, load balancing in MoE is a double-edged sword. Overly pursuing balance can hinder the model's expert specialization and expressive capacity, while neglecting it can lead to issues of training stability and resource wastage. Although current efforts have attempted to address this challenge by incorporating Expert-Level and Device-Level auxiliary loss to encourage tokens to be evenly distributed among experts across multiple devices, these strategies can still lead to training instability and often neglect the influence of model performance. Therefore, future work should focus on developing more effective strategies that ensure balanced utilization of experts and model training stability. Techniques such as adaptive load balancing, dynamic expert capacity adjustment, regularization methods and innovative gating algorithms could be explored.
    \item \textbf{Decentralized:} Decentralized MoE is an exciting but but their potential in combination with other technology paradigms remains underexplored, particularly when combining the principles of decentralization in blockchain with specialized AI models. While this combination holds potential, it also introduces a set of unique challenges that need to be addressed:
    \begin{itemize}[label=$\circ$]
    \item \textbf{High Scalability:} Distributing computational tasks across decentralized nodes can have highly heterogeneous GPUs, CPUs, and network bandwidth, ranging from mobile GPUs to mid-range GPUs in PCs or high-performance GPUs. Network performance, CPUs, storage, and other resources also exhibit high heterogeneity which will create load imbalances and network bottlenecks, limiting scalability. Efficient resource allocation is critical to support a large number of computing nodes to participate, resulting in a high acceleration ratio of system performance.
    \item \textbf{Fault Tolerance:} Decentralized nodes can dynamically join and leave the system, which may cause failures in executing specific AI tasks. This requires that a failure in one or a few nodes does not affect system integrity and availability. Howerver, managing the variability of collaboration and synchronization of updates across distributed experts can lead to issues with model quality and fault tolerance.
    \item \textbf{Communication Latency:} Decentralized nodes often connect via the internet, where communication bandwidth is significantly lower than in high-performance computing clusters. This results in substantial communication latency, especially when exchanging large volumes of data between nodes. Therefore, decentralized MoE systems may experience higher latency due to the need for inter-node communication, which may hinder real-time decision-making applications.
    \item \textbf{Security and Privacy Protection:} Decentralized systems are more vulnerable to attacks. Protecting data privacy and ensuring expert integrity without a central control point is challenging. However, existing methods incur excessive computational overhead.
    \item \textbf{Incentive Mechanism:} Incentive mechanism serve as economic catalysts to encourage users to participate in decentralized system, where nodes with larger contributions are rewarded more, preventing malicious nodes from gaining incentives. Compared with previous incentive schemes in other similar scenarios, there are some special challenges and considerations in the mechanisms design. The property of online training indicates that the arrival and departure time is unknown and varies drastically for different node. This online property is in line with asynchronous training but renders previous one-round incentive schemes ineffective. The incentive mechanism should be robust and resilient to some malicious clients which contribute nothing but endeavor to get large paybacks.
    \end{itemize}    
    These challenges require innovative solutions in decentralized AI architectures, consensus algorithms and privacy-preserving techniques. Advances in these areas will be key to making decentralized MoE systems more scalable, efficient, democratization and secure, ensuring they can handle increasingly complex tasks in a distributed environment.

    \item \textbf{Expert Specialization:} Expert specialization refers to a capability where each expert possesses non-overlapping and highly focused knowledge. Research has demonstrated that encouraging experts to concentrate their skills on specific subtasks or domains significantly enhances the performance and generalization ability of MoE models. However, researchers have found that the so-called "experts" in MoE tend to focus on specific types of tokens or shallow-level concepts. For example, some experts may specialize in handling punctuation marks, while others focus on proper nouns or similar elements. At most, they are grammar-level experts. In addition, the researchers conducted multilingual training on the model. Although one might expect each expert to handle a specific language domain, in fact, this is not the case. More specifically, their expertise lies in processing specific tokens within particular contexts, tending to focus on grammar rather than specific domains. The content that experts learn is more detailed than the broader specific domain. Therefore, future studies should focus on investigating novel mechanisms for enhancing the expert specialization for the development of more powerful MoE models. For instance, introducing meta-learning mechanism into the expert networks to enhance their adaptability and improve the model's ability to handle more complex tasks as well as the specialization of the experts.
\end{itemize}

\section{Conclusion} \label{sect:s6}
In this survey, We introduce the theoretical foundations and core design elements of MoE, such as the sparse activation mechanism of expert networks, routing mechanism, load balancing, and extend beyond centralized systems to delve into decentralized systems. we then exploring its vertical industry applications including medical diagnosis, autonomous driving, financial analysis, business analysis, blockchain, and identify critical challenges and promising directions for future investigation. We hope that this survey serves as a valuable reference for researchers and practitioners, tracking the latest research developments and inspiring new ideas in this explosively evolving field.

\reftitle{References}
%=====================================
% References, variant A: external bibliography
%=====================================
%\bibliography{your_external_BibTeX_file}
\bibliography{references.bib}
%=====================================

\end{document}